\documentclass{rob}%

\usepackage{amssymb}
\usepackage{amsmath}
\usepackage[super]{cite}
\usepackage[linesnumbered,ruled,vlined]{algorithm2e}
\usepackage{graphicx}
\usepackage{multirow}

\usepackage{subcaption}
\usepackage{textcomp}
\doi{10.1017/xxxx}
\volume{31}
\issue{5}
\pubyear{2016}
\cpyear{2016}
\endpage{7}

\begin{document}

\title[Robotica (2016)-Cambridge University Press]{Automated Robotic Monitoring and Inspection of Steel Structures and Bridges}

\author{Hung Manh La$\dagger$\thanks{Corresponding author. E-mail: hla@unr.edu}, Tran Hiep Dinh $\ddagger$, Nhan Huu Pham$\dagger$, Quang Phuc Ha $\ddagger$, Anh Quyen Pham$\dagger$}
\affil{$\dagger$Advanced Robotics and Automation (ARA) Lab, Department of Computer Science and Engineering, University of Nevada, Reno, USA\\
$\ddagger$School of Electrical, Mechanical and Mechatronic Systems, University of Technology Sydney, Australia}

\ADaccepted{MONTH DAY, YEAR. First published online: MONTH DAY, YEAR}

\maketitle

\begin{summary}
This paper presents visual and 3D structure inspection for steel structures and bridges using a developed climbing robot. The robot can move freely on a steel surface, carry sensors, collect data and then send to the ground station in real time for monitoring as well as further processing. Steel surface image stitching and 3D map building are conducted to provide a current condition of the structure. Also, a computer vision-based method is implemented to detect surface defects on stitched images. The effectiveness of the  climbing robot's inspection is tested in multiple circumstances to ensure strong steel adhesion and successful data collection. The detection method was also successfully evaluated on various test images, where steel cracks could be automatically identified, without the requirement of some heuristic reasoning.
\end{summary}


\section{Motivation and Background}
\label{sec:Cmpss}
\begin{figure}[h]
\centerline{\includegraphics[width=0.8\linewidth]{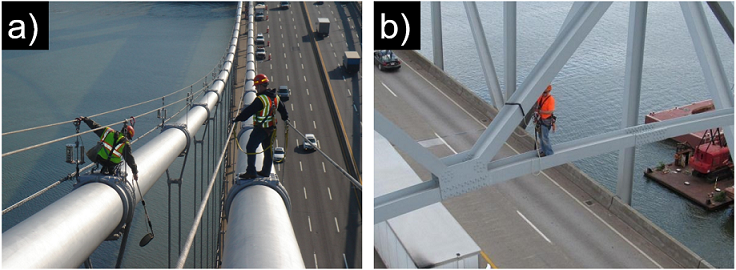}}
    \caption{Dangerous bridge inspection scenes, source: stantec.com.}
    \label{bridgeinspection}
\end{figure}

Steel structures and steel bridges, constituting a major part in civil infrastructure, require adequate maintenance and health monitoring. In the US, more than fifty thousand steel bridges are either deficient or functionally obsolete\cite{FHWA} which present likely a growing threat to people's safety. Collapse of numerous bridges recorded over past 15 years has shown significant impact on the safety of all travelers. For instance, the Minneapolis I-35W Bridge in Minnesota, U.S.A collapsed in 2007\cite{MnDot} due to undersized gusset plates, increased concrete surfacing load, and weight of construction supplies/equipment. This accident along with others have demanded more frequent and effective bridge inspection and maintenance. Currently, steel infrastructure inspection activities require great amount of human effort along with expensive and specialized equipment. Most steel bridge maintenance tasks are manually performed by using visual inspection or hammer tapping and chain dragging for delamination and corrosion detection, which are very time consuming. Moreover, it is difficult and dangerous for inspectors to climb up or hang on cables of large bridges with high structures, as shown in Figure \ref{bridgeinspection}. In addition, reports from visual inspection may vary between inspectors so the structural health condition may not be assessed precisely. Therefore, there should be a suitable approach to provide consistent and accurate reports on steel bridge conditions along with high efficiency and safety assurance in the task execution.

There have recently been an increased number of studies related to utilizing advanced technologies for bridge inspection and maintenance. In H.M. La et al. \cite{LaFR17,La1,La2,La3,LaIROS, LaCASE}, an autonomous robotic system integrated with advanced non-destructive evaluation (NDE) sensors was developed for high-efficiency bridge deck inspection and evaluation while results on real-world  bridge deck crack inspection were reported in  R.S. Lim et al. \cite{Lim1,Lim2}. B. Li et al. \cite{BLi} also utilized the NDE technique to perform automatic inspection on bridge deck and record the health condition of a bridge. For concrete bridges, NDE sensor- integrated robotic systems were deployed by N. Gucunski et al. \cite{Gucunski1,Gucunski2,Gucunski3} to automate the bridge deck inspection. On the other hand, F. Xu et al. \cite{FXu} introduced the design and experiments of a wheel-based cable inspection robotic system, consisting of charge-coupled device (CCD) cameras for the visual inspection purpose. A similar robot \cite{KCho} enables effective visual inspection of the cable on suspension bridges. Moreover, there were several initial implementations of climbing robots for inspection of built infrastructure including steel bridges. A. Mazumdar et al. \cite{Mazu} proposed a legged robot that can move across a steel structure for inspection. Strong permanent magnets embedded in each foot allow the robot to hang from a steel ceiling powerlessly while the attractive force is modulated by tilting the foot against the steel surface. R. Wang, et al. \cite{RWang} developed a robot with magnetic wheels, that is capable of carrying a Giant Magneto Resistive sensor array for crack and corrosion detection. Another bridge inspection method \cite{QLiu,YLiu} featuring a wall-climbing robot with negative pressure adhesion mechanism is used to collect crack images with a high-resolution camera so that a crack can be extracted and analyzed precisely. Based on the attraction force created by permanent magnets, A. Leibbrandt et al. \cite{ALeib} and H. Leon-Rodriguez et al. \cite{HLeon} developed two different wall-climbing robots carrying NDE devices for detection of welding defects, cracks, corrosion testing that can be used to inspect oil tanks or steel bridges. A. San-Millan \cite{AMillan} presented the development of a tele-operated wall climbing robot equipped with various testing probes and cameras for different inspection tasks. D. Zhu et al. \cite{Zhu} used a magnetic wall-climbing robot capable of navigating on steel structures, measuring structural vibrations, processing measurement data and wirelessly exchanging information to investigate field performance of flexure-based mobile sensing nodes to be able to identify minor structural damage, illustrating a high sensitivity in damage detection. Along with the development of climbing robots, computer-vision based methods for crack detection of concrete have been extensively researched and gradually employed in the field to assist with the reduction of costs and hazards of manual data collection \cite{Koch}. Previously, J. K. Oh et al. \cite{Oh} introduced a machine vision system where potential cracks can be identified for bridge inspection by subtracting median filtered images of defects from the original ones. Edge detection algorithms were also employed by R. S. Adhikari et al. \cite{Adhikari} to obtain the skeleton of cracks and compute the corresponding descriptors, such as branch points, length and width profile. R. G. Lins et al. \cite{Lins} recently presented a crack detection algorithm by combining the RGB color model with a particle filter to approximate the probability distribution by a weight sample set. The common point of these techniques is the user input requirement to adjust the filter setting for the effectiveness of the proposed approach. However, this may limit the generic application of an automatic inspection system. Our contribution in this paper is a new crack detection technique based on a non-parametric peak detection algorithm where a threshold to isolate crack from the image background is automatically calculated. The proposed technique is only based on a basic knowledge that crack pixels are darker than their surroundings.

In this paper, the inspection technique using visual and 3D images coupled with histogram thresholding and image segmentation is applied to our developed climbing robot. Here, the robot is able to strongly adhere to and freely move on a steel surface to collect images from visual as well as 3D cameras. The captured images from the robotic visual camera are then stitched together to provide a whole image of the steel surface for ease of inspection. Then an automatic detection algorithm is developed to detect defects from steel surface's images. Moreover, a 3D map is built from 3D point cloud data in order to assist with robot navigation as well as inspection of bridge structures. With an advanced mechanical design, the robot is able to carry a payload up to approximately 7 kg while still adhering to both inclined and upside-down surfaces. The robot can also transit from one surface to another with up to $90^{\circ}$ change in orientation. Overall, steel surface image stitching and 3D structure mapping are performed from collected data to provide condition assessments for the steel structures.

The rest of paper is organized as follows. Section \ref{Sec:Design} describes the overall robotic system design. Section \ref{Sec:Data} presents data collection and image processing techniques. Section \ref{Sec:Result} shows experimental results to verify the system design and data processing. Finally, Section \ref{Sec:Conc} gives the conclusion and future work.

\section{Overall Design of the Inspection Robotic System}\label{Sec:Design}

\subsection{Mechanical Design}

\begin{figure}
\centerline{\includegraphics[width=0.5\linewidth]{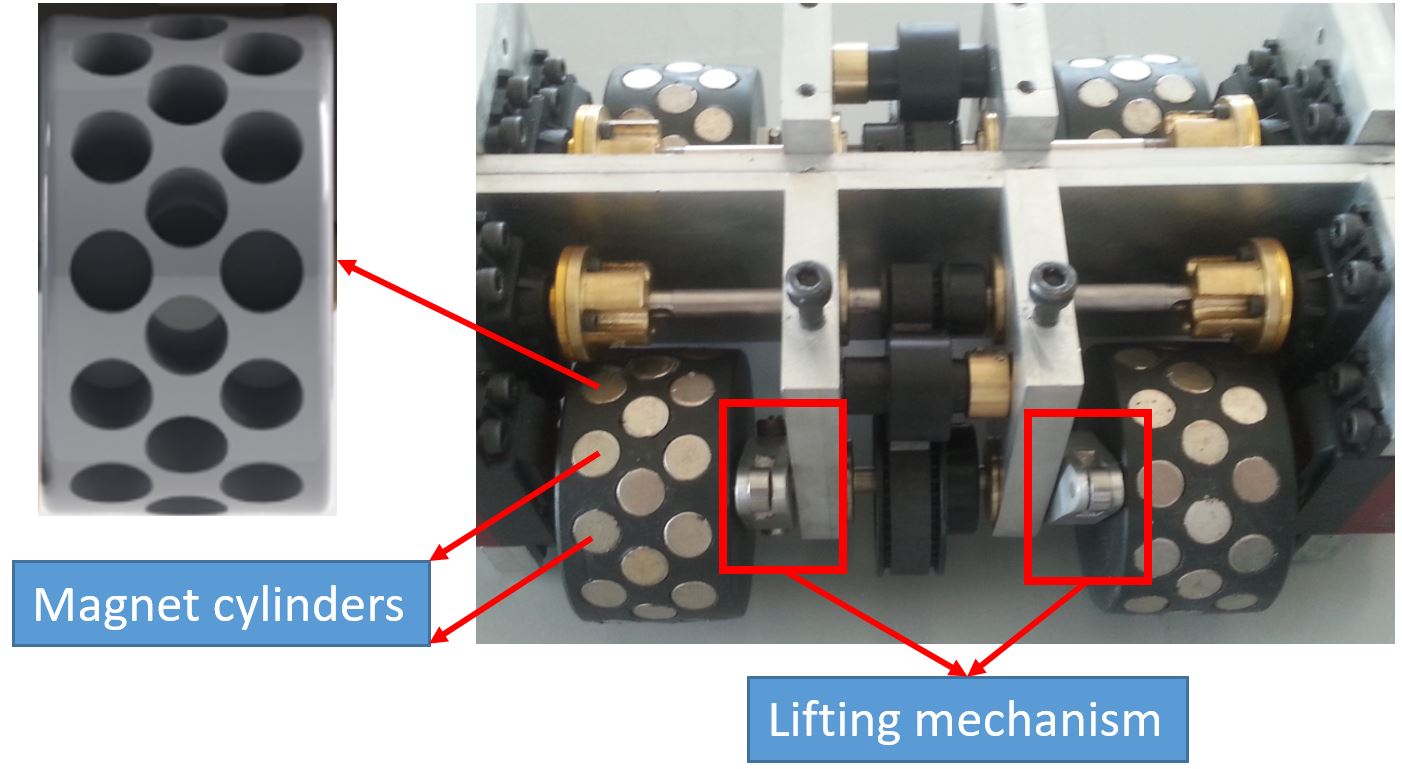}}%
\caption{Embedded magnet cylinders and lifting mechanism.}
\label{fig:magnet}
\end{figure}

\begin{figure}
\centerline{\includegraphics[width=\linewidth]{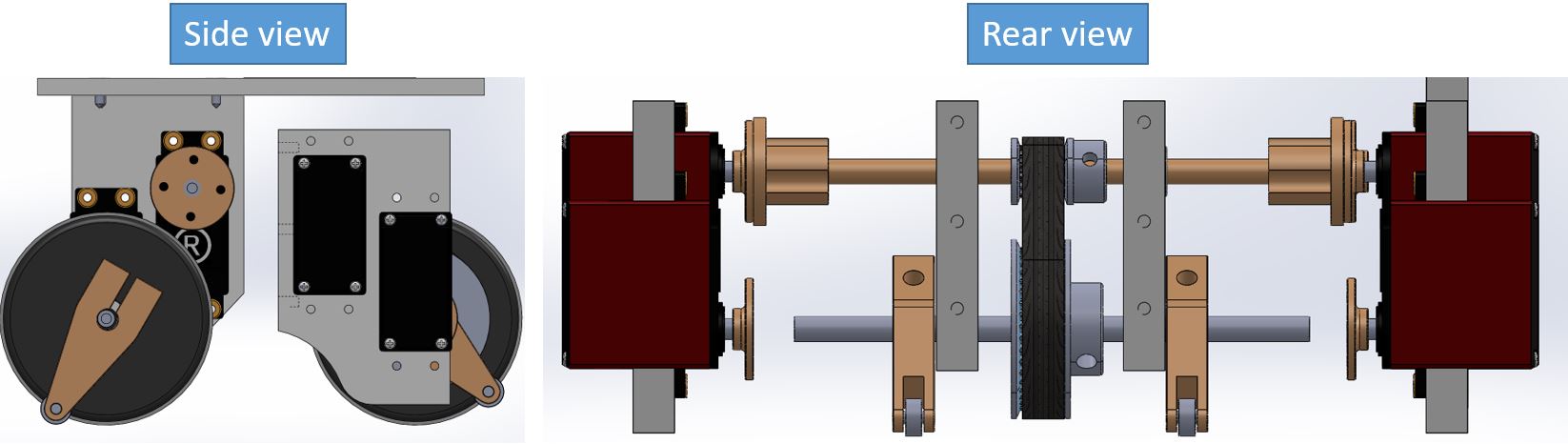}}%
\caption{Lifting mechanism design.}
\label{fig:liftDesign}
\end{figure}

\begin{figure}
\centerline{\includegraphics[width=\linewidth]{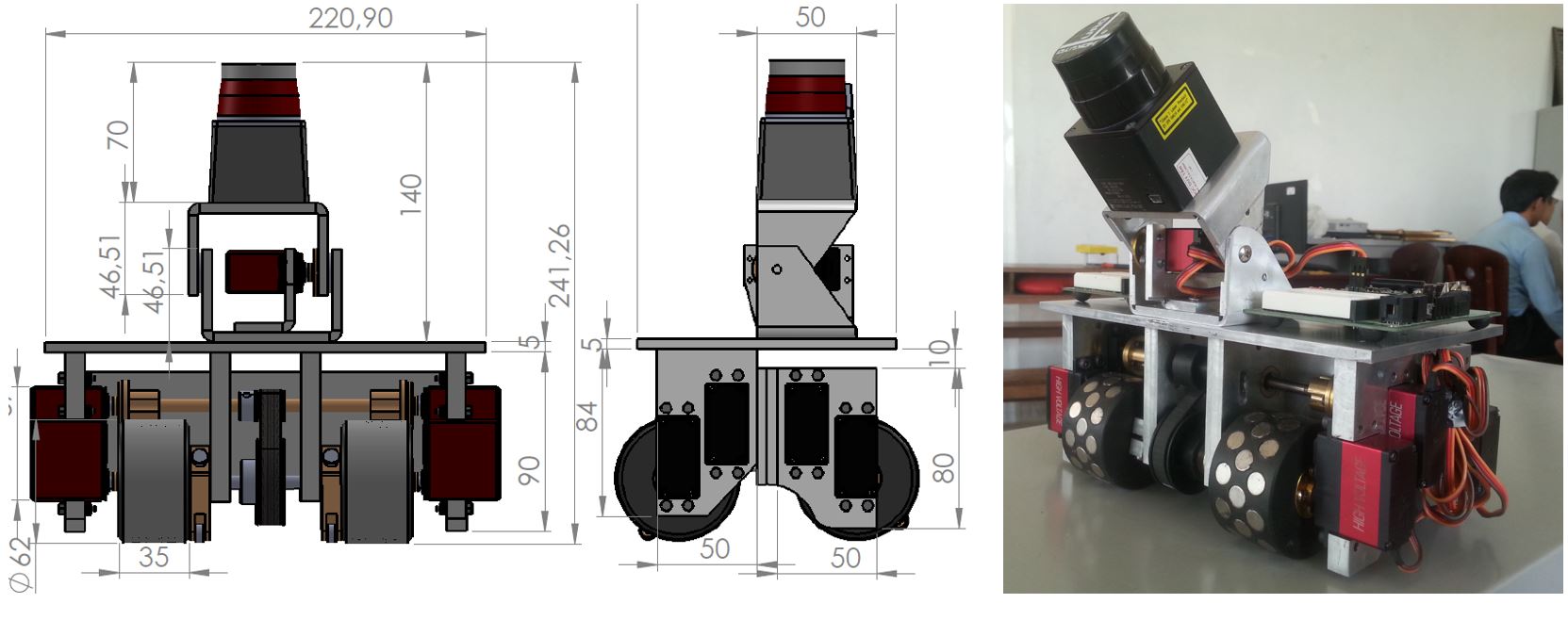}}%
\caption{Robot design and initial prototype.}
\label{fig:3dview}
\end{figure}

A robot design with four motorized wheels is proposed, taking the advantage of permanent magnets for adhesion force creation. This allows the robot to adhere to steel surfaces without consuming any power. In addition, cylinder shaped magnets are used for convenience and flexibility in adjusting the attraction force. Moreover, eight high torque servo motors are utilized to drive four wheels for robot navigation and four shafts for robot lifting. A total of 36 Neodymium magnet cylinders\cite{KJ} embedded in each wheel can create magnetic force up to 14 Kg/wheel. To enhance the friction with steel surface each wheel is covered by a thin layer of cloth. However, this also causes the magnetic force created is reduced to approximately 6 Kg/wheel. Hence, four modified servo motors with a higher torque (3.3 Nm) are used to drive the robot\cite{JR}. Additionally, a special mechanism has been designed in order to lift either the front or rear wheels off the ground if the robot is stuck on rough terrains. Figure \ref{fig:magnet} depicts the magnet cylinders installation and the lifting mechanism mentioned while Figure \ref{fig:liftDesign} and \ref{fig:3dview} illustrate the lifting mechanism design and the overall robot design with an initial robot prototype, respectively.

\begin{figure}
\centerline{\includegraphics[width=0.8\linewidth]{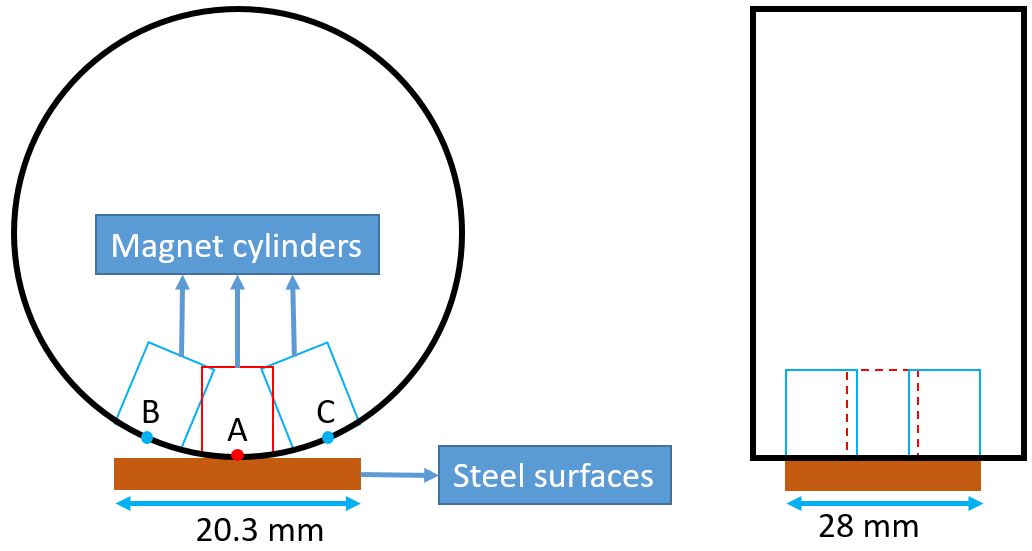}}%
\caption{Magnetic force analysis under limited steel surfaces.}
\label{fig:limitedsurface}
\end{figure}

While moving on steel surfaces, it is required to maintain stability of the robot to overcome sliding and turn-over failures. It has been shown that in order to avoid these failures, the magnetic force created by all wheels should satisfy \cite{Pham_2016}
\begin{equation} \label{eq:fmag} 
	{F}_{mag} > max\Big\{\dfrac{P\sin\alpha}{\mu}+P\cos\alpha;2\dfrac{Pd}{L}\Big\},
\end{equation}
where $F_{mag}$ is the total magnetic force created by all wheels, $\alpha$ is the degree of inclination of the steel surface ($0\le \alpha \le 90^{\circ}$), $\mu$ is the frictional coefficient between the wheel cover and steel surface, $P$ is the robot's weight, $d$ is the distance between the center of mass to the surface, and $L$ is the distance between the front and rear wheels. Condition (\ref{eq:fmag}) is essential for selecting appropriate robot design parameters. In detail, beside adjusting the total magnetic force, one can satisfy (\ref{eq:fmag}) by reducing the robot's weight or alternating the design to decrease ratio $\dfrac{d}{L}$. In this design, in order to maintain (\ref{eq:fmag}) the steel surface area underneath each robot's wheel must be at least $20.3mm \times 28 mm$ as shown in Figure \ref{fig:limitedsurface}. Therefore, in the presence of bolts, if the contact areas are large enough, the robot will be able to pass through. Otherwise, the robot needs to transition to other surface to move forward. More details of the robot's mechanical design can be referred to \cite{Pham_2016}.

\subsection{Robot Control}

\subsubsection{System Integration}

Regarding sensing devices, the robot is equipped with multiple imaging sensors: two video cameras for image capturing and video streaming, and a time-of-flight (ToF) camera for capturing 3D data. The USB camera at the back of the robot is installed facing downward  in order to take images of steel surfaces and feed to the image stitching unit. Besides, the ToF camera is placed so that 3D images received from the camera can assist with robot navigation as well as building 3D structures.

Apart from cameras, eight Hall Effect sensors are used to detect the presence of a magnetic field. Two sensors are mounted next to each other and close to one wheel. By observing that magnet cylinders inside each wheel will move when the robot moves, we can extract the velocity and traveling distance of each wheel after combining the data collected from these two sensors.

\begin{figure}
\centerline{\includegraphics[width=0.6\linewidth]{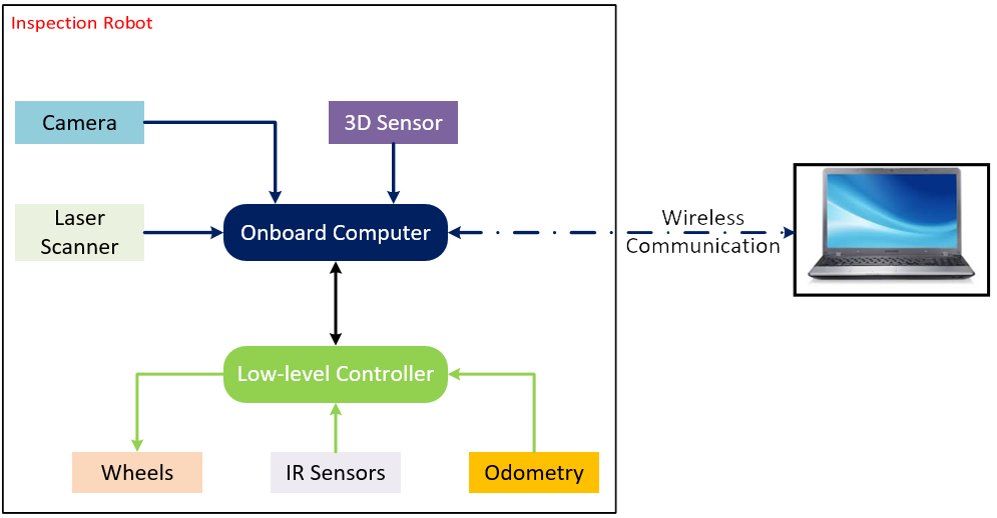}}%
\caption{Robotic system architecture.}
\label{fig:structure}
\end{figure}

\begin{figure}
\centerline{\includegraphics[width=\linewidth]{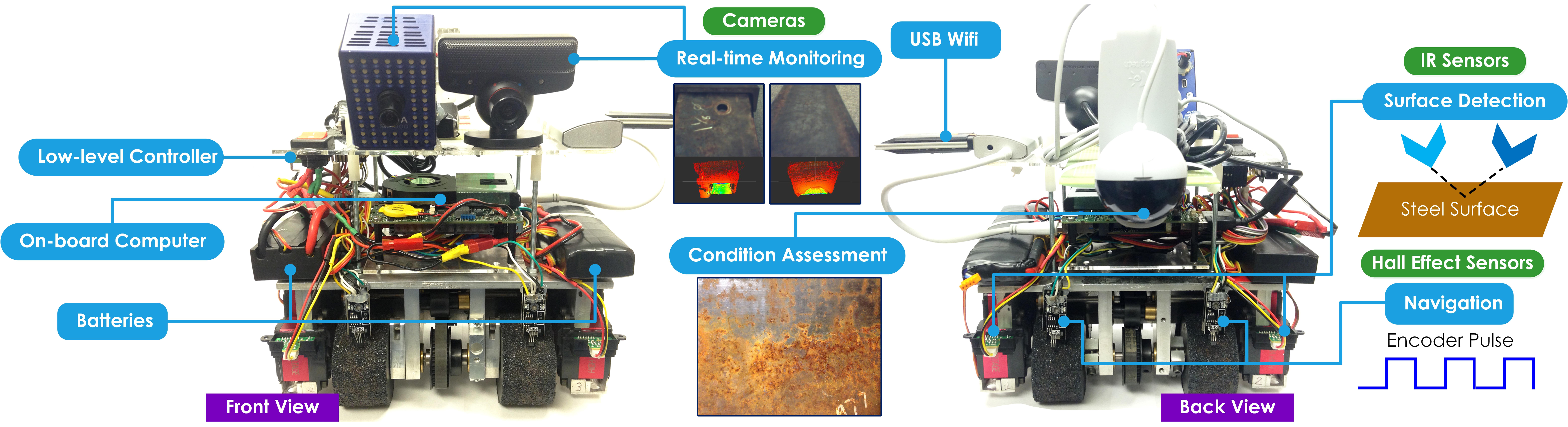}}%
\caption{Robot prototype with integrated sensors.}
\label{fig:sensors}
\end{figure}

Moreover, the robot has four IR range sensors mounted at four corners of the robot, which can detect whether there exists a surface underneath. Consequently, an edge avoidance algorithm can be implemented using this input to make sure that the robot can safely travel on steel surfaces.
The robot is controlled by a microcontroller unit (MCU) handling low-level tasks and a more powerful on-board computer for complex processing and communication with the ground station. The low-level controller has the capability of receiving commands from the on-board computer via serial connections, including motion control and sensors data acquisition signals. The on-board computer is an Intel NUC Core i5 computer responsible for capturing video camera and 3D camera images, then sending them to the ground station over wireless LAN connection for data post processing and logging. It also executes the edge avoidance algorithm with sensors data received from the low-level controller to ensure safe traveling on steel surfaces. The whole robot is powered by two batteries with different voltage levels. One $12V$ battery powers the on-board computer and cameras while another $7.4V$ battery supplies to all motors.
Overall, the structure of the system is shown in Figure \ref{fig:structure} while the robot with fully installed sensors and other components is depicted in Figure \ref{fig:sensors}. More details of the robot control can be referred to \cite{Pham_2016}.

\subsubsection{Robot Navigation}

While moving on steel surfaces, there exists a circumstance that the robot moves far away toward the edge of the surface, and may fall off. Therefore, an algorithm using input from IR range sensors is incorporated to prevent this incident. Let us denote $r\_cal_{i} (i=1:4)$  the calibrated ranges before the  robot starts moving, $r_i (i=1:4)$  as IR sensor reading corresponding to $sensor_i$ and $d_i (i=1:4)$ the travel distances calculated from Hall Effect sensors. Given a predefined threshold $\epsilon$, when sensor reading is out of the range $[r\_cal_{i} - \epsilon ; r\_cal_{i} + \epsilon]$, the robot considers that there is no surface below $sensor_i$. The algorithm then adjusts the robot's heading to avoid falling out of the surface. The summary of the edge avoidance algorithm for safe navigation is presented in Algorithm \ref{algo:Edge}.
\begin{algorithm}
\DontPrintSemicolon 
\KwIn{$(r\_cal_{1},r\_cal_{2},r\_cal_{3},r\_cal_{4})$, $(r_1,r_2,r_3,r_4)$, $\epsilon$, $(d_1,d_2,d_3,d_4)$}

\For {i=1:4}
{
	\uIf {only one $(r_i) \notin [r\_cal_{i} - \epsilon ; r\_cal_{i} + \epsilon]$}
	{
		\uIf {i== front right IR sensor}
		{
			Stop\;
			Go backward with a distance of $5 cm$ $(\Delta d_i \approx 3)$\;
			Rotate left when travel distance of either right wheel reach $3 cm$ $(\Delta d_i \approx 2)$ \;
			Keep moving\;
		}
		Check other sensors and take similar actions\;
	}
	\Else 
	{
		stop and wait for commands\;
	}
}
\caption{{\sc Edge Avoidance.}}
\label{algo:Edge}
\end{algorithm}

It should be noted that the velocity and distance calculated from Hall Effect sensor reading also provide information to implement automatic operations of images capturing  so that we can periodically collect images then stitch them together before applying crack detection algorithms. To ensure that two images can be stitched together, the robot need to stop after traveling a particular distance so that the two consecutive images are overlapped with a minimum percentage.

\section{Data Collection and Processing}\label{Sec:Data}

\subsection{Image Stitching}

\begin{figure}
\centerline{\includegraphics[width=0.6\linewidth]{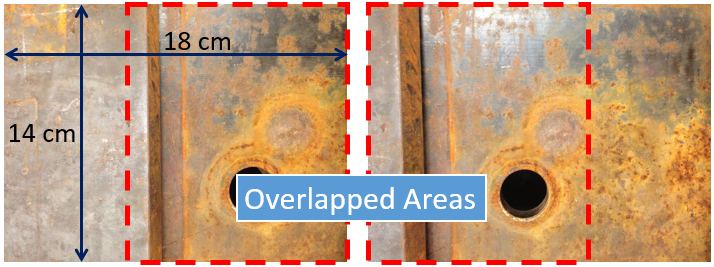}}%
\caption{Overlapped images used for stitching.}
\label{fig:overlap}
\end{figure}

In order to enhance steel surface inspection, we combine images captured from the camera at the back, considered as a case of image stitching. Captured images are saved to the memory so that they can be processed later when the robot has finished its operation.
In this technique, it is required that two consecutively taken images are at least $30\%$ overlapped as shown in Figure \ref{fig:overlap}. The camera can cover an area of 18cm $\times$ 14cm which means the robot should stop and capture images every 12 cm.

Camera motion is estimated incrementally using sparse feature-matching and image-to-image matching procedures \cite{DFor,MBrown}. The image is processed left-to-right to find the location of overlapping inside the other image. We can enhance the process by providing an initial estimation of the overlapping area between two consecutive images using robot motion estimation since the camera is fixed on the robot. Then we can use an appearance-based template matching technique to achieve finer results of the camera motion. If two images have a significant difference in brightness, a technique is applied to determine a new exposure value of the combined image. If the  arriving pixel is significantly brighter than the existing corresponding pixel, we replace the old pixel with the new one. A threshold value $\tau$ is used to consider whether it is brighter or not. The exposure compensation and blending technique applied to two consecutive images is then described by the following intensity:
\begin{equation} \label{eq:exposure} 
I(x,y)= \left\{\begin{array}{ll}
		I_2(x,y)~\text{if}~~\tau I_2(x,y)>I_1(x,y) \\
		I_1(x,y)~\text{otherwise}.
	\end{array}\right.
\end{equation}
Gaps in the region formed by the pixels to be used from the new image are filled by using a 2D median filter. As a result, the completeness of the shadow removal region is maintained. 

\begin{figure}
\centerline{\includegraphics[width=0.6\linewidth]{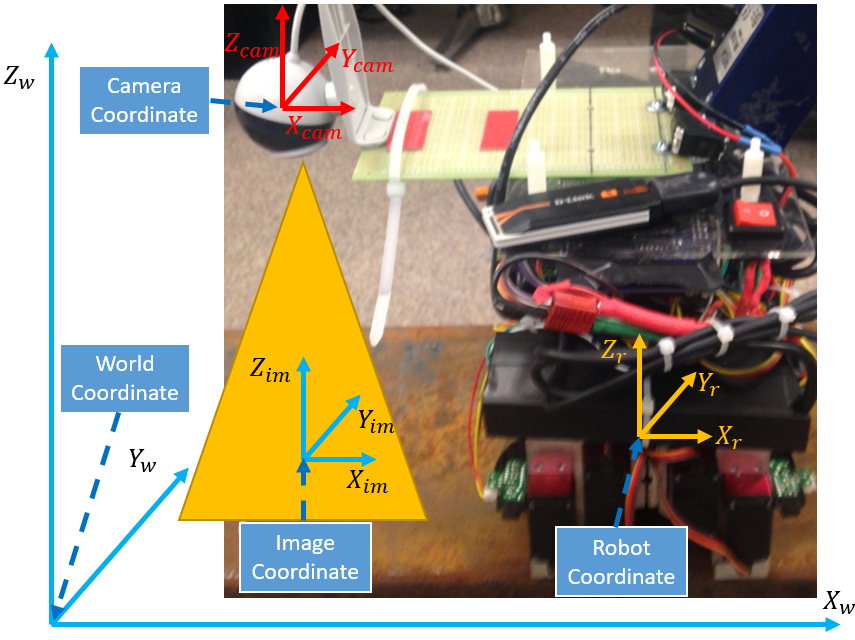}}%
\caption{Relationship between multiple frames.}
\label{fig:transform}
\end{figure}

While stitching all of the images, we can also transform the stitched image's coordinate to the world's coordinate. Denote $(X_{im},Y_{im},Z_{im})$ the coordinates in the image frame, $(X_{cam},Y_{cam},Z_{cam})$  coordinates in the camera frame, $(X_r,Y_r,Z_r)$  coordinates of the robot frame and $(X_w,Y_w,Z_w)$ coordinates of the world fixed frame, as shown in Figure \ref{fig:transform}.

The following series of transformation should be done to convert coordinates in the image frame to the world frame
\begin{equation*}
(X_{im},Y_{im},Z_{im})\xrightarrow{T_{ic}}(X_{cam},Y_{cam},Z_{cam})\xrightarrow{T_{cr}}(X_r,Y_r,Z_r) \xrightarrow{T_{rw}}(X_w,Y_w,Z_w)
\end{equation*}
where $T_{ij}$ are the transform matrix from frame $i$ to frame $j$. Since the camera location is fixed on the robot, $T_{ic}$ and $T_{cr}$ can be easily obtained by measuring the distance between the camera and the steel surface and the robot's center of mass while $T_{rw}$ can be extracted from odometry.

The resulting stitched image can be really useful in real life application. Instead of viewing hundreds or thousands of steel surface images, inspectors can have an overall image of the whole steel surface that the robot has traveled. Hence, it improves the efficiency of visual inspection of a bridge.

\subsection{3D Construction}

\begin{figure}
\centerline{\includegraphics[width=\linewidth]{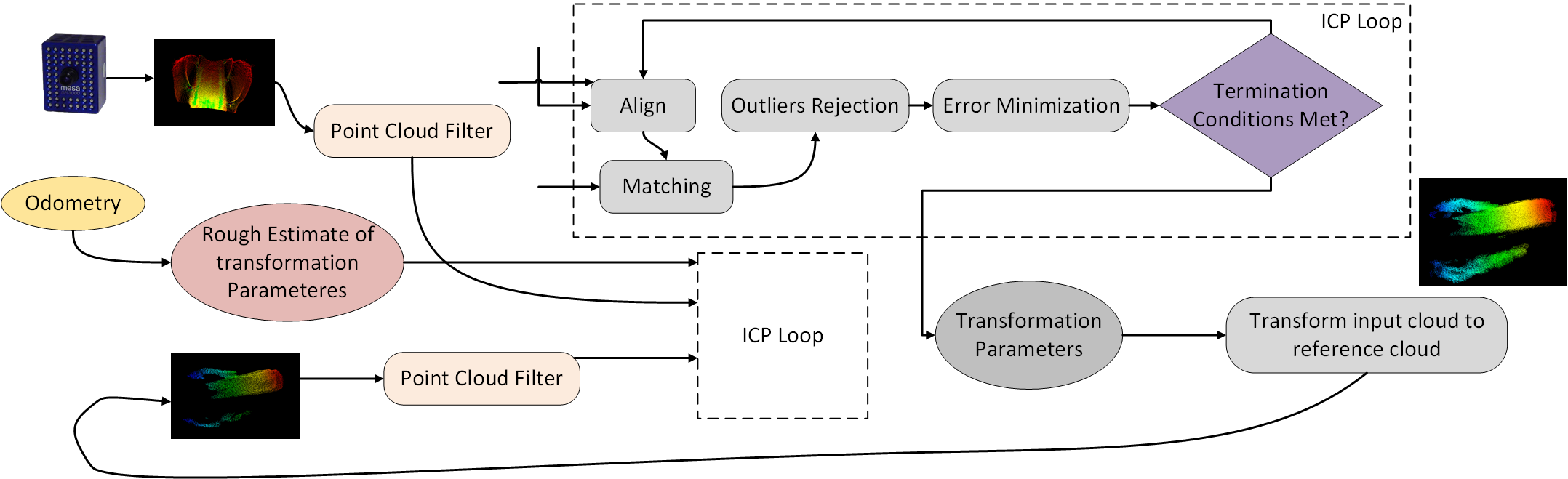}}%
\caption{3D map construction process employing ICP algorithm.}
\label{fig:icp}
\end{figure}

The concept of 3D construction or 3D registration is based on the Iterative Closest Point (ICP) algorithm introduced in Besl et al. \cite{PBesl} and Y. Chen\cite{YChen}. The algorithm has been used to construct 3D models of objects in robotic applications including mapping and localization. Here, the ICP goal is to find the transformation parameters (rotation and translation) that align an input point cloud to a reference point cloud. Those transformation parameters are presented in the coordinate frame of the reference point cloud. Figure \ref{fig:icp} shows the process of 3D registration including the ICP algorithm, which consists of these steps: \cite{DHolz}

\begin{enumerate}
 \item Selection: Input point clouds captured from a time-of-flight (ToF) camera are pre-sampled for a higher efficiency.

 \item Matching: Odometry data can be used to estimate correspondences between the points in the sub-sampled point clouds, considered as the initial alignment. After that, we apply feature matching using Fast Point Feature Histograms (FPFH) \cite{Rusu2009} on two roughly aligned images to obtain finer transformation.

 \item Rejection: Filtering the correspondences to reduce the number of outliers, multiple points with the same corresponding point are rejected.

 \item Alignment: Computing assessment criteria, normally point-to-point or point-to-plane error metrics, then minimizing them to find an optimal transformation.
\end{enumerate}

The algorithm stops at one of these cases:
\begin{enumerate}
 \item The error metrics decrease to within or remain constant below a threshold.
 
 \item The algorithm does not converge after a given number of iterations.
 
 \item Transformation parameters do not vary or are out of bound.
\end{enumerate}

\subsection{Steel Crack Detection}

\begin{figure}[htp]
\centerline{\includegraphics[width=\linewidth]{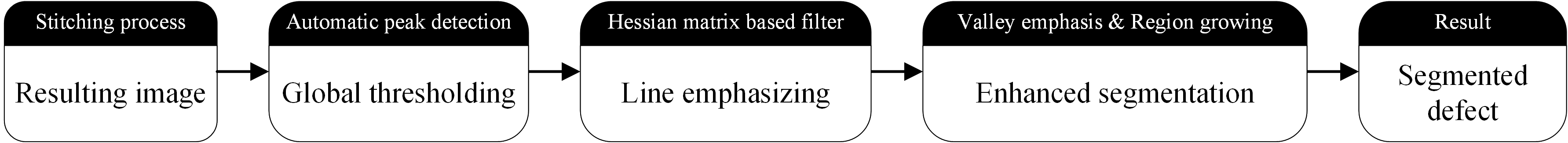}}%
\caption{Proposed approach for steel crack detection.}
\label{fig:scd}
\end{figure}

In robotic applications, many vision-based approaches have been proposed to deal with robotic color tracking and image segmentation \cite{Yu_AT, Hiep, Hiep2,Koch} and employed to solve the defect detection problem \cite{NG_HF, Dirami, Yuan}. To be applied in a fully automatic process, the proposed approach must be able to handle the input data with limited supervision and manual adjustment.In this work, a hybrid method is proposed to solve the mentioned problem by combining our automatic peak detection algorithm with image stitching and 3D registration. At this stage, the stitched images from data collected by the cameras are further processed to automatically detect a corrosion or crack on the steel coating surface. The potential structural or surface defect is firstly isolated from the image background by using a global threshold, which is obtained from our automatic peak detection algorithm. A Hessian matrix based filter is then employed to eliminate the blob- or sheet-like noise and to emphasize the defect nature \cite{YSato, YFuji}. To avoid the disconnectivity of identified cracks or the 
misdetection of thin coating surfaces, the region growing algorithm, see, e.g. \cite{Zhao}, is applied to enhance the segmentation performance. Figure \ref{fig:scd} illustrates the processing steps for our proposed approach.

\subsubsection{Automatic peak detection algorithm}

\begin{figure}
\centerline{\includegraphics[width=\linewidth]{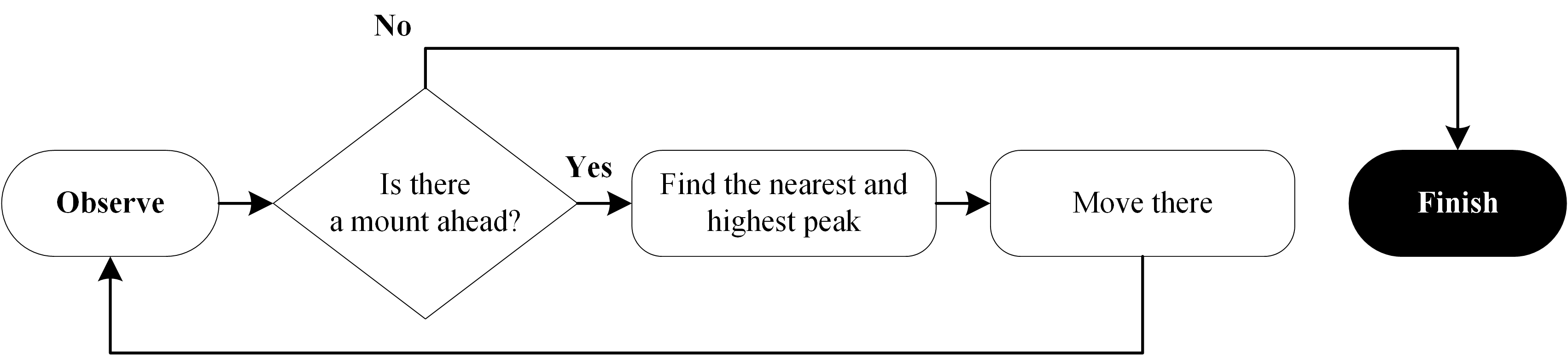}}%
\caption{Strategy for a lost mountains explorer.}
\label{fig:lostmountain}
\end{figure}

The proposed algorithm used in this work is based on perceptions of a mountain explorer whereby observation and planning remain the key factors to be considered when getting lost in mountain exploration. In this circumstance, a feasible solution is to repeat the process of finding a high location to observe and identifying a specific landmark to head to as well as planning for the next travel segment until the explorer finally gets back on track. To illustrate this strategic planning, a flowchart is described in Figure \ref{fig:lostmountain}. Therein, two main steps are involved, namely, searching for observing location and for the highest peak in the closest distance.

The gray-scale histogram of the resulting image is firstly smoothed using a moving average filter with the width of kernel equal to 3, taking into account the previous, current and next intensity level. This filter is chosen for our approach because of its compactness and robustness in eliminating random noise while retaining significant peaks of the histogram. Let $h(i)$ be the pixels number at intensity level $i$ for $i=1,2,\ldots,L+1$ where $L$ is the maximum intensity level. After applying the moving average filter, the pixels number at intensity level $i$ is determined as:
\begin{equation} \label{eq:pixelNum} 
h(i)=\dfrac{1}{3}[h(i-1)+h(i)+h(i+1)].
\end{equation}

An initial peak of the smoothed histogram is identified if its intensity value is greater than that of its two nearest neighbors. The significant peak detection strategy is then built based on these initial ones. Therefore, initial peaks can be considered as a draft plan and stored in a cluster vector $\delta$, where each element must meet the following criteria:
\begin{equation} \label{eq:pixelNum2} 
\left\{\begin{array}{ll}
h(\delta(k))>h(\delta(k)-1)\\
h(\delta(k))>h(\delta(k)+1).
\end{array}\right.
\end{equation}

\paragraph{Observing location}
Let consider the smoothed histogram as the mount to be explored. A point on the intensity axis could then be considered as the observing location $\alpha(m)$ if the following condition is fulfilled:
\begin{equation} \label{eq:observingLoc} 
\alpha(m)<g(\delta(k))-L(\delta(k)),
\end{equation}
where $L(\delta(k))$ is a dynamic offset distance from the current peak to the observing location and dependent on the draft plan, $g(\delta(k))$ is the intensity level at the $k^{th}$ initial peak and $p_D$ is the number of possible dominant peaks, $1\le m\le p_D$. The offset distance is defined as:
\begin{equation} \label{eq:offsetDist} 
L(\delta(k))= \left\{\begin{array}{ll}
		\dfrac{h(\delta(k))[g(\delta(k+1))-g(\delta(k))]}{|h(\delta(k+1))-h(\delta(k))|}\indent\text{if}\indent h(\delta(k+1))\neq h(\delta(k))\\
		\dfrac{h(\delta(k))[g(\delta(k+1))-g(\delta(k))]}{|\dfrac{k+1}{k}h(\delta(k+1))-h(\delta(k))|}\indent\text{if}\indent h(\delta(k+1))=h(\delta(k)).
	\end{array}\right.
\end{equation}

It can be seen that this offset distance $L(\delta(k))$ is determined based on their own pixel number, and the two adjacent intensity levels correspondingly, as illustrated in Figure \ref{fig:Offset}. Based on the height difference and distance between two adjacent peaks, a higher peak could always be identified from the calculated observing location. In the implementation phase, the observing location is set at $g(\delta(k))-L(\delta(k))$, hence:
\begin{figure}
		\centering
		\begin{subfigure}{.5\textwidth}
			\centering
			\includegraphics[width=200pt]{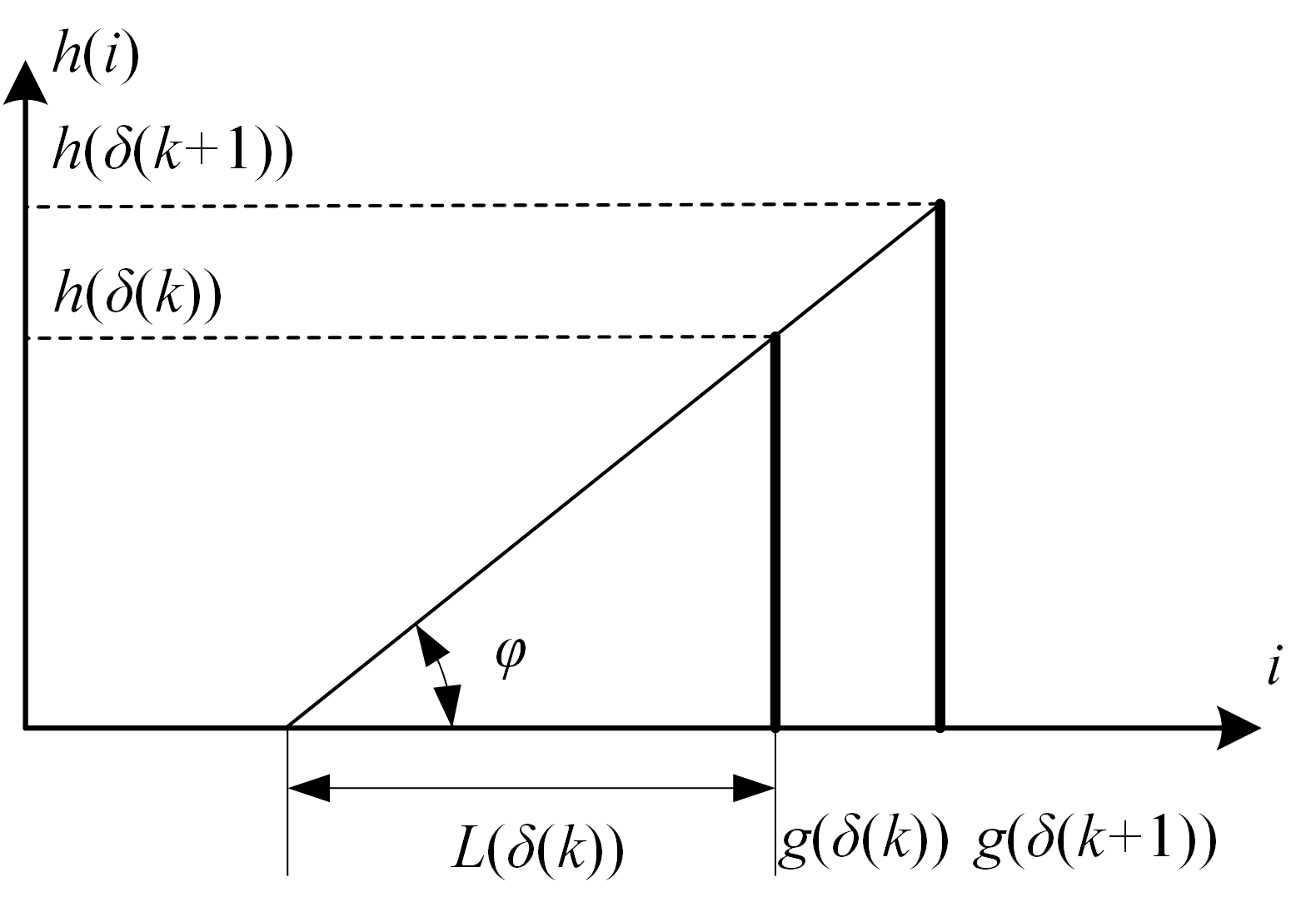}
			\caption{}
			\label{fig:2a}
		\end{subfigure}%
		\begin{subfigure}{.5\textwidth}
			\centering
			\includegraphics[width=200pt]{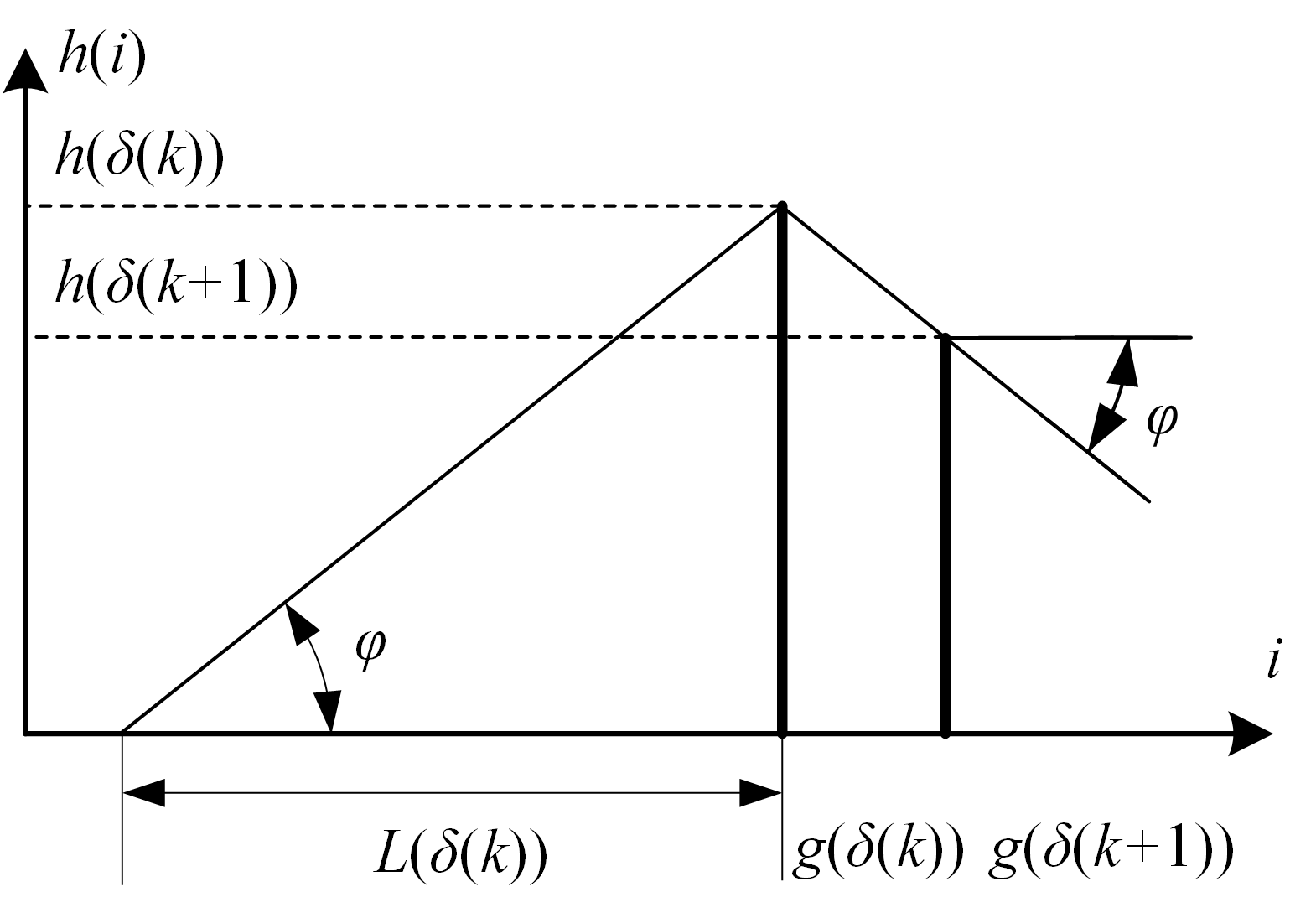}
			\caption{}
			\label{fig:2b}
		\end{subfigure}	
		\caption{Offset distance determination:\\ (a) $g(\delta(k+1))>g(\delta(k))$, and (b) $g(\delta(k))>g(\delta(k+1)).$ }
		\label{fig:Offset}
	\end{figure}
\begin{equation} \label{eq:observingLoc2} 
\alpha(m)=g(\delta(k))-L(\delta(k)),
\end{equation}
and the observing location is then considered as the highest in the neighborhood to maintain an unobstructed view. For the $k^{th}$ detected peak in the cluster vector, the crossover index is proposed as follows:
\begin{equation} \label{eq:possibility} 
\theta(\delta(k))=\dfrac{d(\delta(k))}{L(\delta(k))},
\end{equation}
where
\begin{equation*}
d(\delta(k))=h(\delta(k))-\dfrac{min\{h(\delta(k)),h(\delta(k+1))\}}{2}.
\end{equation*}

\paragraph{Highest peak in the closest distance}
A peak is considered as a nearest prominent peak, if it is high enough to be``observed" from the current observing location and to ``block'' the view to the next peak. This statement could be mathematically described as follows:
\begin{equation} \label{eq:peak} 
\left\{\begin{array}{ll}
		\theta(\delta(k))>\theta(\delta(k+1))\\
		\theta(\delta(k))>\theta(\delta(k-1)).
	\end{array}\right.
\end{equation}

\begin{figure}
\centerline{\includegraphics[width=\linewidth]{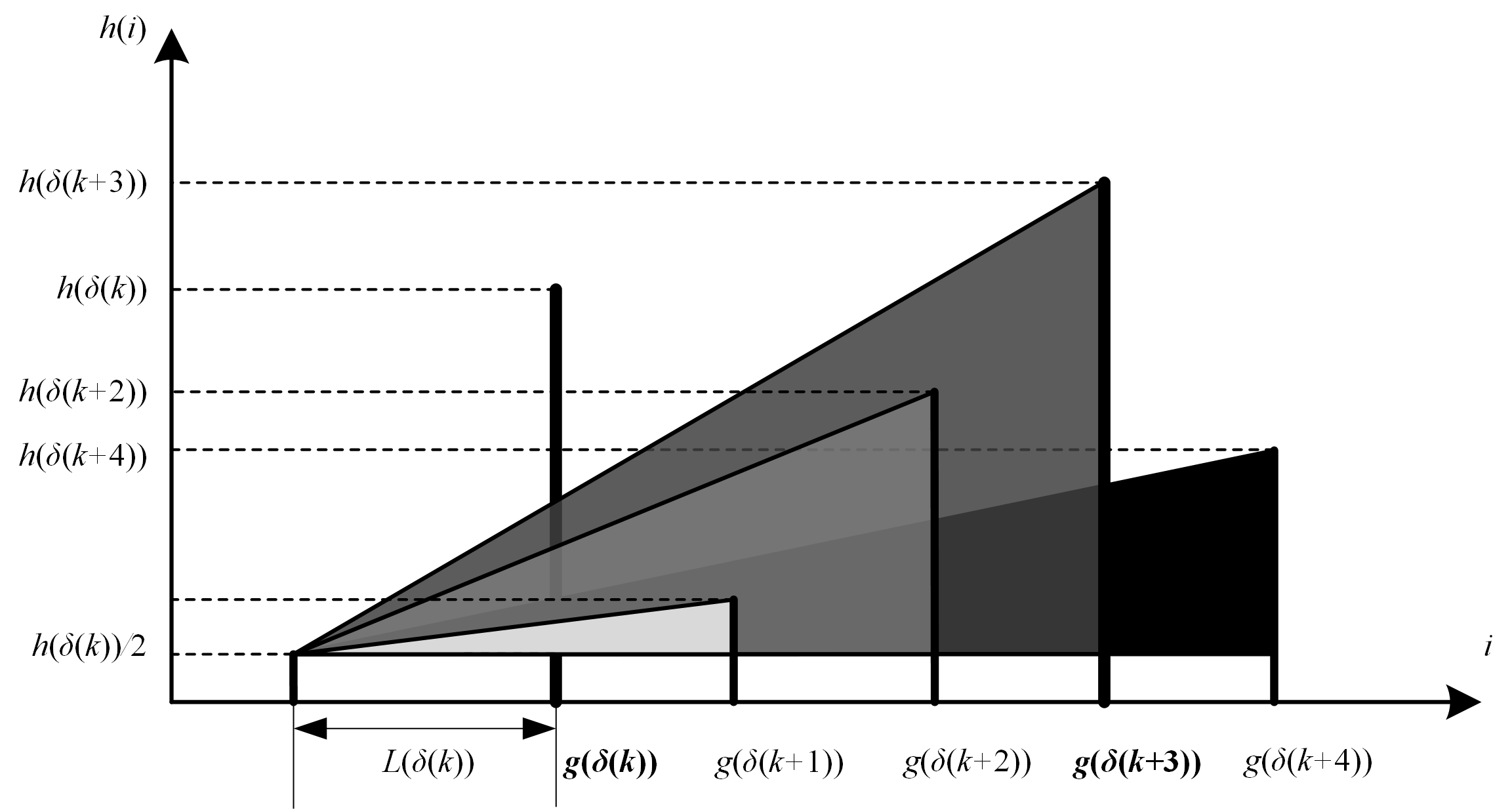}}%
\caption{Illustration of searching mechanism.}
\label{fig:SearchMec}
\end{figure}

The searching mechanism of our algorithm is illustrated in Figure \ref{fig:SearchMec}. Firstly, $g(\delta(k))$ is identified as a dominant peak. The new location for observation will be updated to $g(\delta(k))-L(\delta(k))$. The crossover index of $g(\delta(k))$ is reset while this characteristic is re-calculated for $g(\delta(k+1)),g(\delta(k+2)),\ldots$ based on the updated observing location. A following dominant peak is then determined at $g(\delta(k+3))$. Similarly, when a new observing location is calculated, the crossover index of the next peak is consequently updated according to Eqs. (\ref{eq:offsetDist}-\ref{eq:possibility}) until the cluster vector is completely checked. Algorithm \ref{algo:Peak} illustrates the implementation of the proposed algorithm pseudo code. The identified peaks are then used to partially segment the cracks and rust from the background.


Compared to some recent histogram-based peak detection algorithms \cite{Sezan, YuanB}, the major advantage of the proposed approach is that no prior knowledge or information is required to detect all main peaks of the image histogram. While user-specified parameters are compulsory in methods based on the peak sensitivity and peaks number \cite{YuanB}, the offset distance and crossover index in our technique can be calculated and updated automatically based on the location of detected initial peaks. Using Intel(R) Core(TM) i5-5300U CPU @2.30 GHz with 64 bit Windows 7, the average computation time for detecting peaks of a 8 bit histogram was approximately 12 ms for our algorithm and 15 ms for the peak detection signal method \cite{YuanB} with the main reason being the smaller number of loops used in the searching process.

\begin{algorithm}
\DontPrintSemicolon 

m=1;n=1;found = 0\;
\For {k=n:length($\delta$)}
{
	// calculate peak's crossover index\;
	\For {j=(n+1):(k-1)}
	{
   	\uIf {there is a dominant peak at level j}
	  {
		  $n = j + 1$\;
		  $found = 1$\;
		  // save the position of dominant peak in $\beta$\;
		  $\beta(m) = \delta(j)$\;
		  // update position of observing location\;
		  $m = m + 1$\;
		  $\alpha(m) = g(\delta(j)) - L(\delta(j))$\;
		  // break out of \textbf{for} loop\;
		  break\;
		}
   }
   \uIf {found}
   {
   	fprintf(``Dominant peak at position \%d'',$j$)\;
   	fprintf(``New observe position at \%d'',$\alpha(m)$)\;
   }
   \Else
   {
   	  fprintf(``No dominant peak at this step'')\;
   }
}
\caption{{\sc Automatic Peak Detection.}}
\label{algo:Peak}
\end{algorithm}

\subsubsection{Line emphasizing}

We implement a line emphasis filter, proposed by Sato et al. \cite{YSato}  and Fujita et al. \cite{YFuji} based on the Hessian matrix to remove blob-like and sheet-like noise and feature the line structure corresponding to a crack. The Hessian matrix of an image $I(\textbf{x})$, where $\textbf{x}=(x,y)$ is given by:
\begin{equation} \label{eq:Hmatrix1} 
\nabla^{2} I(x)= \begin{bmatrix}
 I_{xx}(\textbf{x}) &I_{xy}(\textbf{x}) \\
 I_{yx}(\textbf{x}) &I_{yy}(\textbf{x})
 \end{bmatrix},
\end{equation}
where
\begin{equation} \label{eq:Hmatrix2} 
\left\{\begin{array}{ll}
		I_{xx}(\textbf{x})=\dfrac{\partial^2}{\partial x^2}I(\textbf{x})\\
		I_{xy}(\textbf{x})=\dfrac{\partial^2}{\partial x\partial y}I(\textbf{x})\\
		I_{yy}(\textbf{x})=\dfrac{\partial^2}{\partial y^2}I(\textbf{x})\\
		I_{yx}(\textbf{x})=\dfrac{\partial^2}{\partial y\partial x}I(\textbf{x}).
	\end{array}\right.
\end{equation}

The eigenvalues  $\lambda_1 (\textbf{x})$, $\lambda_2 (\textbf{x})$ of $\nabla^{2} I(\textbf{x})$, where $\lambda_1\ge \lambda_2$, are adopted to acquire a generalized measure of similarity to a line structure as:
\begin{equation} \label{eq:eigen} 
\lambda_{12}=\left\{\begin{array}{ll}
		|\lambda_2|\Big(1+\dfrac{\lambda_1}{|\lambda_2|} \Big)=|\lambda_2|+\lambda_1~\textbf{if}~\lambda_2 \le \lambda_1 \le 0\\
		|\lambda_2|\Big(1-\mu\dfrac{\lambda_1}{|\lambda_2|} \Big)=|\lambda_2|-\mu\lambda_1~\textbf{if}~\lambda_2 < 0 < \lambda_1 < \dfrac{|\lambda_2|}{\mu}\\
		0~~\text{otherwise},
	\end{array}\right.
\end{equation}
where $0 < \mu \le 1$.

The line-, blob- or sheet structure in the image could be expressed by combining two eigenvalues $\lambda_1$ and $\lambda_2$ as described in Tab. \ref{tbl:engenShape}.

\begin{table*}
\tbl{Combination of eigenvalues and corresponding shape structures}{
\begin{tabular}{lll} \colrule
 Relationships between eigenvectors & Structure \\ 
 \botrule
 $|\lambda_1|\ge |\lambda_2| \approx 0$ & Line \\
 $|\lambda_1|\approx |\lambda_2| \ge 0$ & Blob \\
 $|\lambda_1|\approx |\lambda_2| \approx 0$ & Sheet \\
 \botrule
\end{tabular}}
\label{tbl:engenShape}
\end{table*}

In order to recover line structures of various widths, the partial second derivatives of the image $I(\textbf{x})$ in Eq. (\ref{eq:Hmatrix2}) can be combined with the Gaussian convolution, for example,
\begin{equation} \label{eq:GausCon} 
I_{xx}(\textbf{x};\sigma_f)=\Big\{\dfrac{\partial^2}{\partial x^2}G(\textbf{x};\sigma_f) \Big\}*I(\textbf{x}),
\end{equation}
where $G(\textbf{x};\sigma_f)$ is an isotropic Gaussian function with standard deviation $\sigma_f$. The maximum among the normalized multiple scales will be selected from the multi-scale integration of the filter responses of a pixel $\textbf{x}$ within a region defined by $R(\textbf{x})$ as:
\begin{equation} \label{eq:FiltRes} 
R(\textbf{x})=\max_{\sigma_i} \sigma_1^2 \lambda_{12}(\textbf{x};\sigma_i),
\end{equation}
where $\sigma_i=s^{i-1}\sigma_1(i=1,2,\ldots,n)$ is a discrete sample of $\sigma_f$, with $\sigma_1$ and $s$ being the minimum scale and scale factor of the sampling interval of $\sigma_f$, respectively.

\subsubsection{Enhanced Segmentation}
The thresholding and emphasizing step can partially extract a defect of steel surfaces from the background of images but may face a difficulty in image segmentation of thin cracks or rusty areas. To overcome this disadvantage, we employ a region growing algorithm based on a dynamic threshold obtained from the Otsu thresholding method \cite{NG_HF} and our peak automatic detection approach. The boundary pixels of each extracted abnormal region are identified as the initial seed points. Let $g_S$ denote the mean intensity value of the concerning region and $g_N$ denote the intensity value of any seed point in the neighborhood, a similarity criterion can be simply judged by the following difference:
\begin{equation}\label{eq:Crit}
e=|g_S-g_N|.
\end{equation}    

\begin{figure}
		\centering
		\begin{subfigure}{.5\textwidth}
			\centering
			\includegraphics[width=200pt]{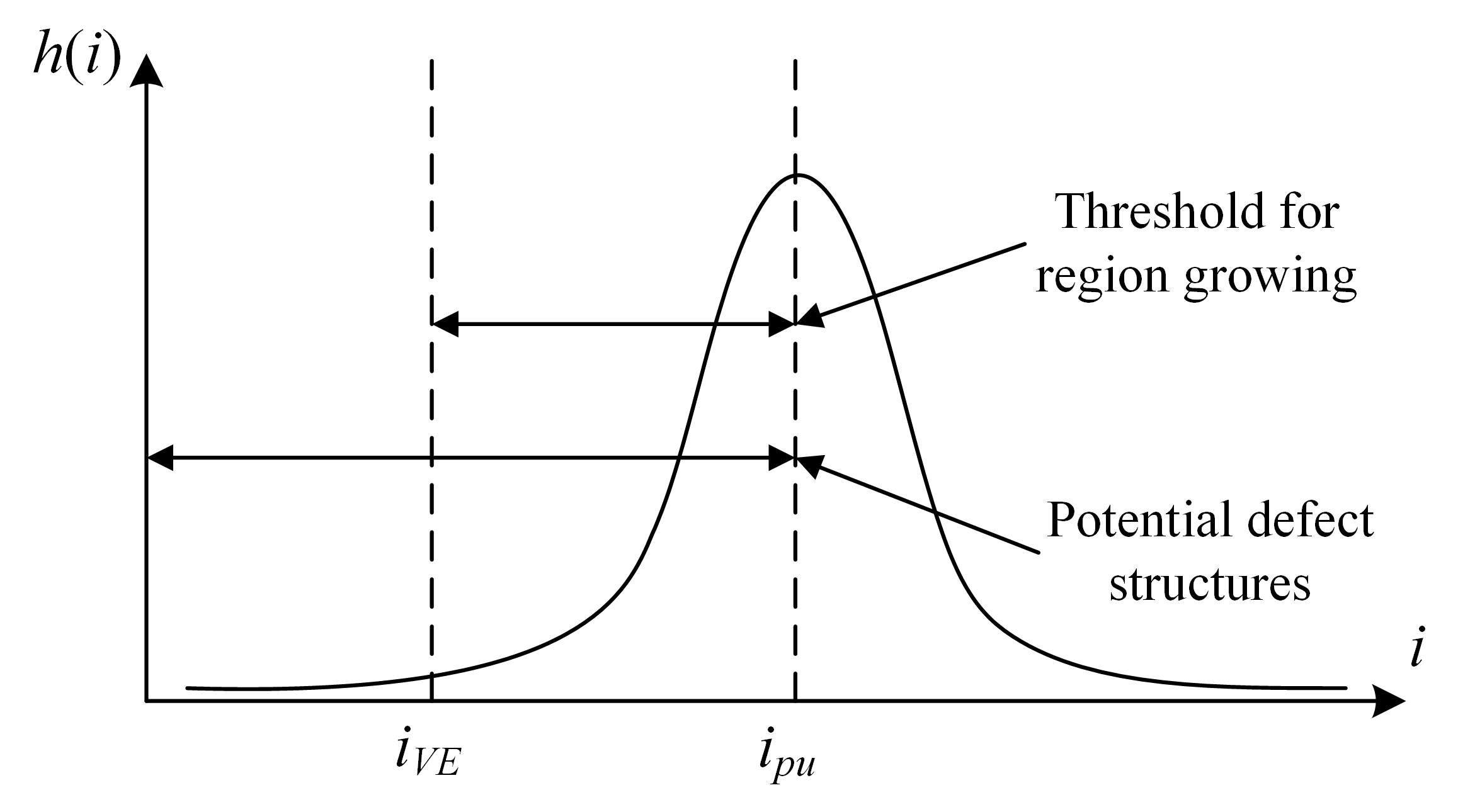}
			\caption{}
			\label{fig:2a}
		\end{subfigure}%
		\begin{subfigure}{.5\textwidth}
			\centering
			\includegraphics[width=200pt]{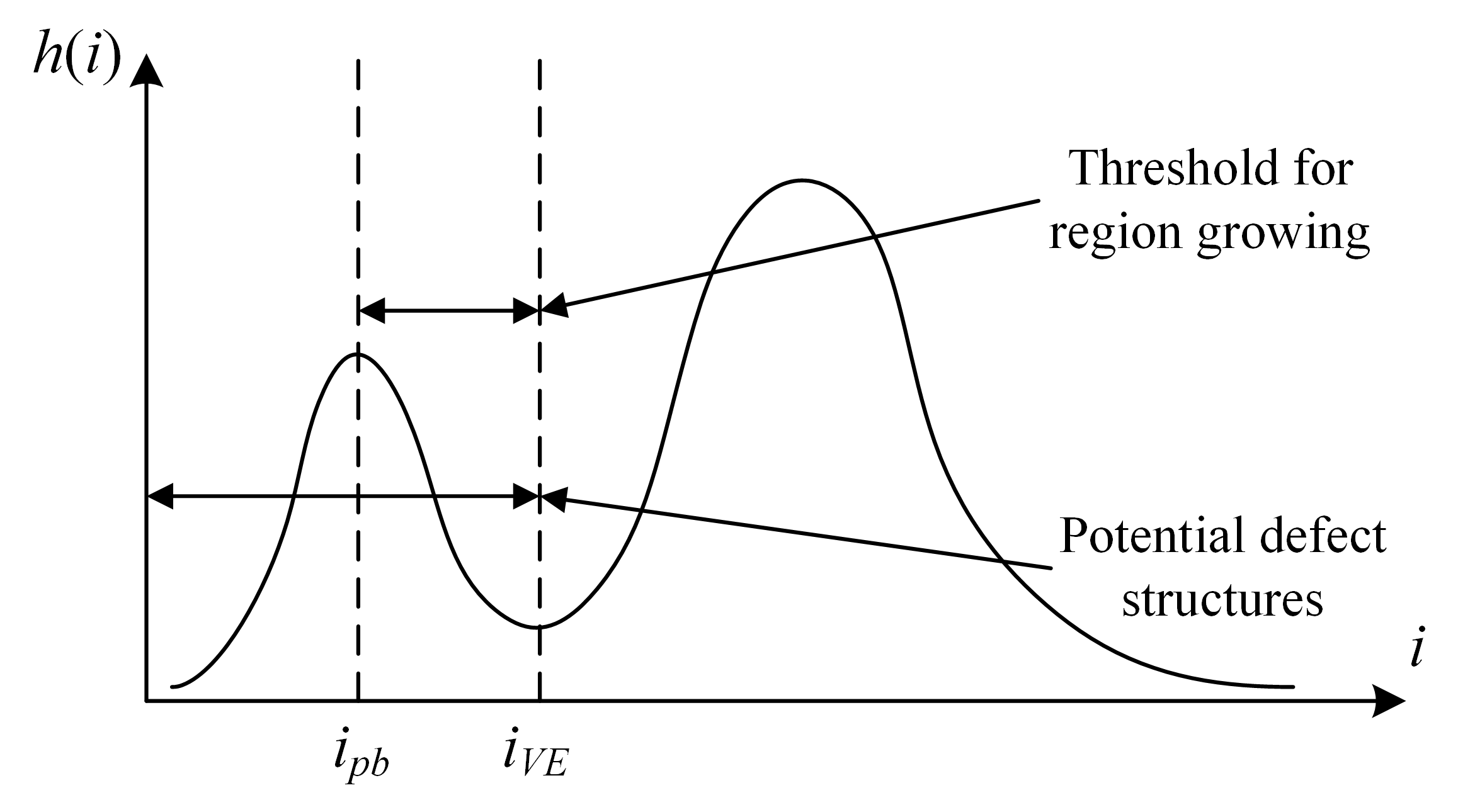}
			\caption{}
			\label{fig:2b}
		\end{subfigure}	
		\caption{Threshold for region growing:\\ (a) Unimodal, and (b) Bimodal. }
		\label{fig:RegGrow}
	\end{figure}

A neighbor pixel can be added in the detected defect region if its intensity difference with the region mean is smaller than a pre-defined threshold. In this paper, the Otsu method in combination with the location of detected peaks is applied to identify the threshold and to determine the valley in the unimodal and bimodal histogram, as illustrated in Figure \ref{fig:RegGrow}. The formulation for the valley-emphasis method is:
\begin{equation}\label{eq:OpThresh}
t^*=\arg\max_{0 \le 1<L}\Big\{(1-p_t)(\omega_1(t)\mu_1^2(t)+\omega_2(t)\mu_2^2(t))\Big\},
\end{equation}
where $t^*$ is the optimal value of the threshold $t$ between the foreground class $C_1=\{0,1,...,t\}$ and the background class $C_2=\{t+1,t+2,...,L\}$ of the processed image, in which $L$ is the number of distinct gray levels. The probabilities of the two classes are respectively,
\begin{equation}
\omega_1(t)=\sum_{i=0}^tp_i\indent\text{and}\indent \omega_2(t)=\sum_{i=t+1}^Lp_i,
\end{equation}
where $p_i=\frac{h(i)}{n}$ is the probability of occurrence of gray-level $i$, $\sum_{i=1}^np_i=1$, and $n$ is the total number of pixels in the concerning image.

Here, by region growing, the detected defect area is expanded to the neighbor pixel by using the similarity criterion via the intensity difference between the histogram peak and the emphasized valley.

\section{Experimental Results and Discussion}\label{Sec:Result}

In order to verify the design and assess the performance of the developed robotic system, both indoor and outdoor experiments were conducted. The indoor test was within the lab environment on small steel bars while the outdoor experiment was on a real-world steel bridge. The ability of climbing, handling navigation difficulties and surface condition assessment were evaluated in a number of scenarios. During the tests, one 2S1P (2 cells) 7.4V 5000 milliampere-hour (mAh) and one 3S1P 11.1V 5000 mAh batteries are used to power the robot. One laptop which can connect to a wireless LAN is used as a ground station. The total weight of the robot is about $6$ Kg while the total magnetic force created is approximately $16$ Kg, satisfying condition (\ref{eq:fmag}).

\begin{figure}
\centerline{\includegraphics[width=\linewidth]{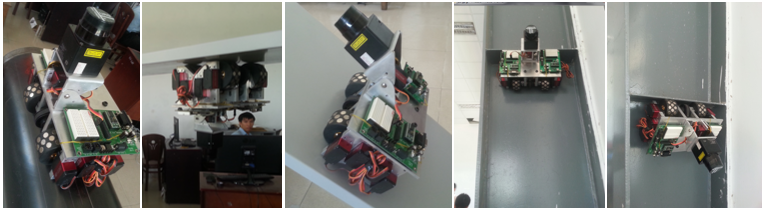}}%
\caption{Adhesion tests on different surfaces under different degrees of inclination.}
\label{fig:adhere}
\end{figure}

\begin{figure}
\centerline{\includegraphics[width=\linewidth]{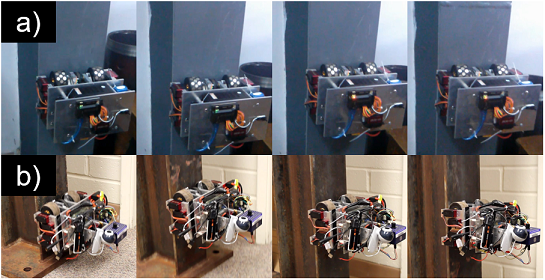}}%
\caption{Robot moves on steel surfaces: (a) Without load, and (b) With full load.}
\label{fig:move}
\end{figure}

Figure \ref{fig:adhere} presents various cases when the robot was placed on steel surfaces under different degrees of inclination to ensure that the magnetic force is strong enough to adhere to steel surfaces when the robot does not move. During the experiments, the climbing capability tests are done on a bridge and on several steel structures nearby with coated or unclean surfaces. Although the surface is curvy, the robots can still adhere tightly to the steel structures. It also shows strong climbing capability on a rusty surface. Moreover, robot is  capable of vertically moving, without carrying or with a load, as shown in Figure \ref{fig:move}.

\begin{figure}
\centerline{\includegraphics[width=\linewidth]{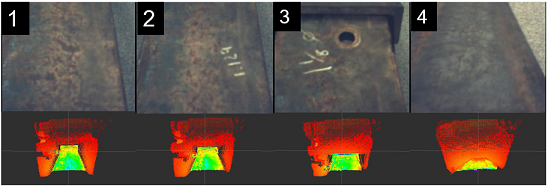}}%
\caption{Visual and 3D images acquired from cameras assist robot navigation and map construction. (Top) Visual image; (Bottom) 3D image of the structure.}
\label{fig:Data}
\end{figure}

\begin{figure}
\centerline{\includegraphics[width=\linewidth]{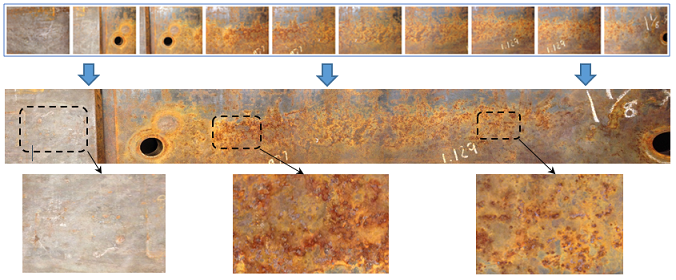}}%
\caption{Images stitching result: (Top) 10 individual images taken by the robot; (Middle) Stitching image result from those 10 individual images; (Bottom) Closer look (zoom-in) at some areas, from left-to-right, showing “good condition”, “serious deteriorated condition”, and “light deteriorated condition” of the steel surface, respectively.}
\label{fig:stitch}
\end{figure}

\begin{figure}
		\centering
		\begin{subfigure}{.5\textwidth}
			\centering
			\includegraphics[width=160pt]{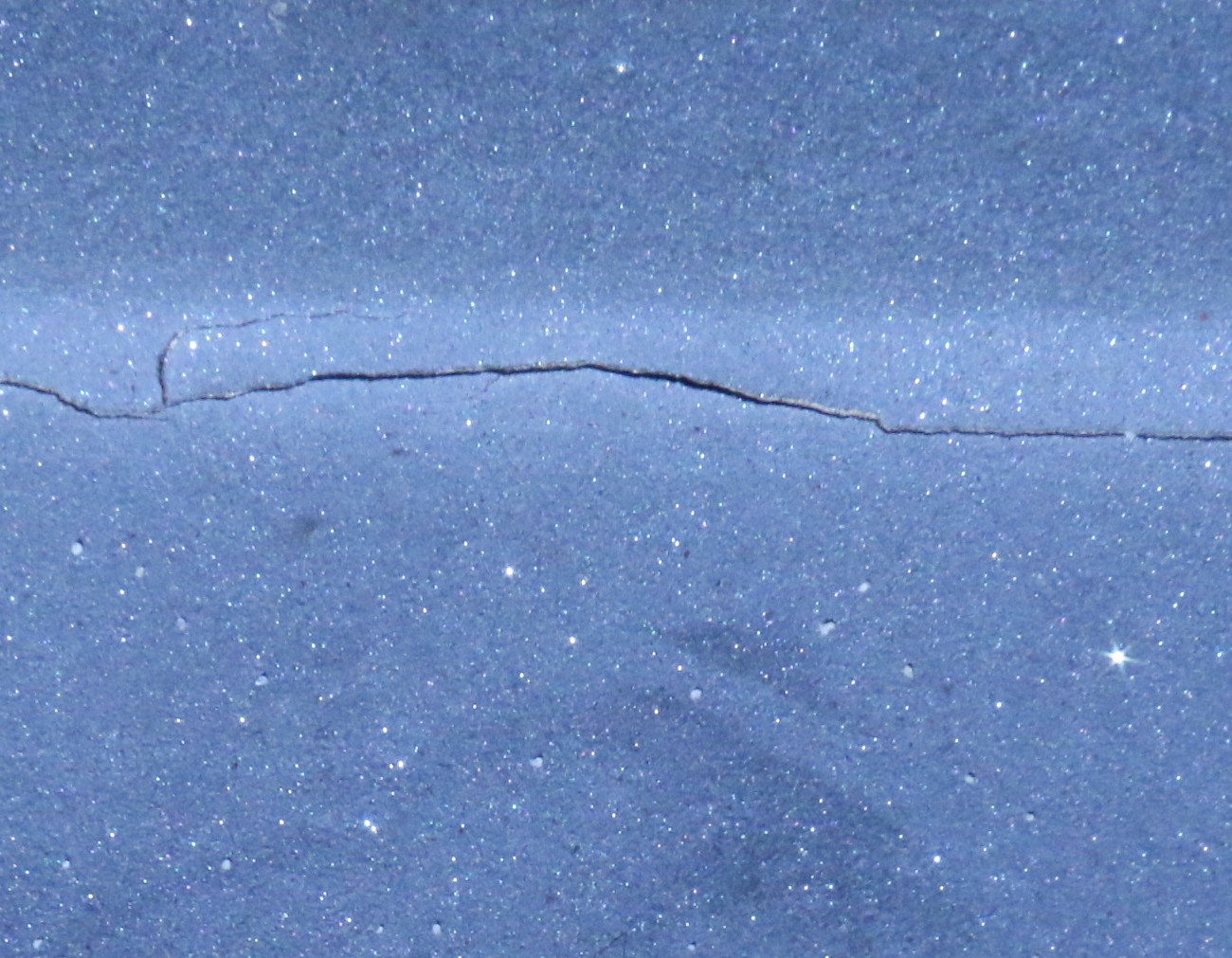}
			\caption{}
			\label{fig:2a}
		\end{subfigure}%
		\begin{subfigure}{.5\textwidth}
			\centering
			\includegraphics[width=160pt]{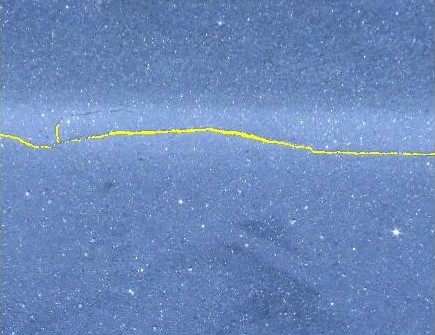}
			\caption{}
			\label{fig:2b}
		\end{subfigure}	
		\begin{subfigure}{.5\textwidth}
					\centering
					\includegraphics[width=200pt]{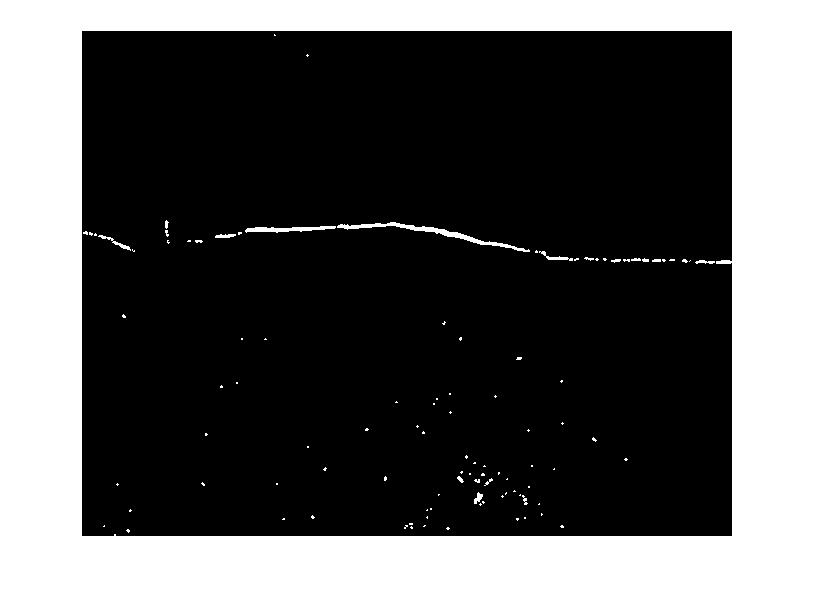}
					\caption{}
					\label{fig:2c}
				\end{subfigure}%
				\begin{subfigure}{.5\textwidth}
					\centering
					\includegraphics[width=200pt]{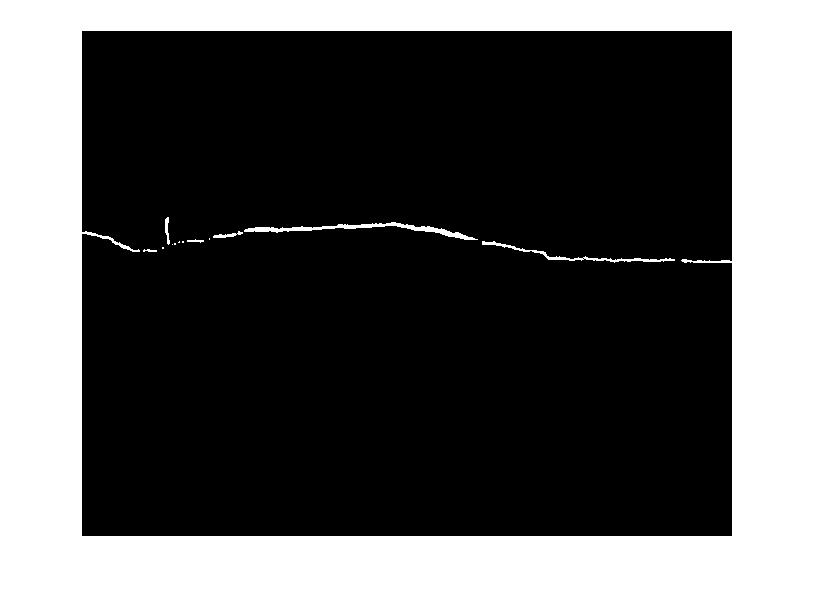}
					\caption{}
					\label{fig:2d}
				\end{subfigure}	
		\caption{Crack detection result:\\ (a) Surface image, (b) Detected crack on surface image, (c) Global thresholding and line emphasizing and (d) Enhanced segmentation.}
		\label{fig:Crack}
	\end{figure}

\begin{figure}
	\centering
	\begin{subfigure}{.5\textwidth}
		\centering
		\includegraphics[width=160pt]{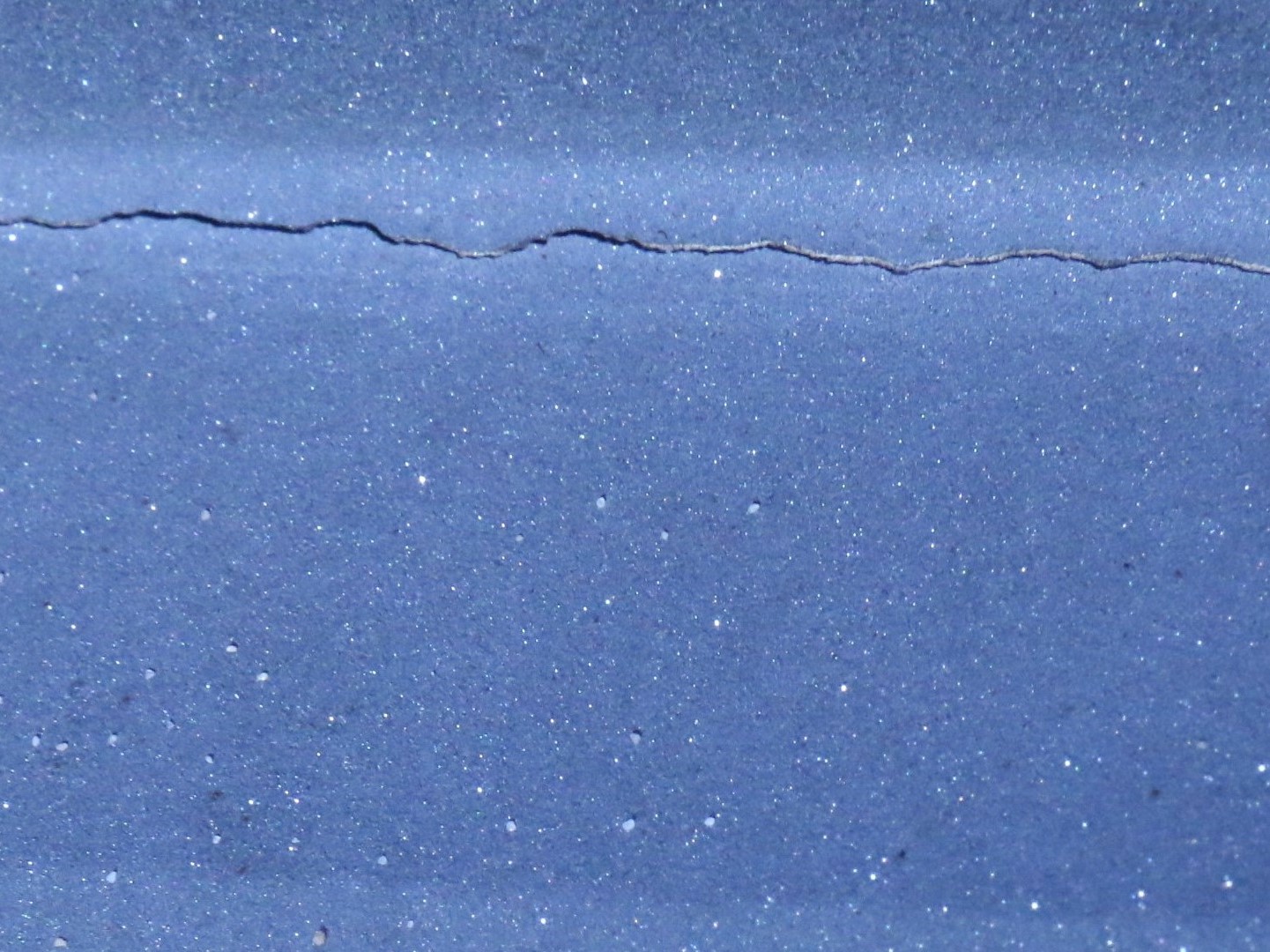}
		\caption{}
	\end{subfigure}%
	\begin{subfigure}{.5\textwidth}
		\centering
		\includegraphics[width=160pt]{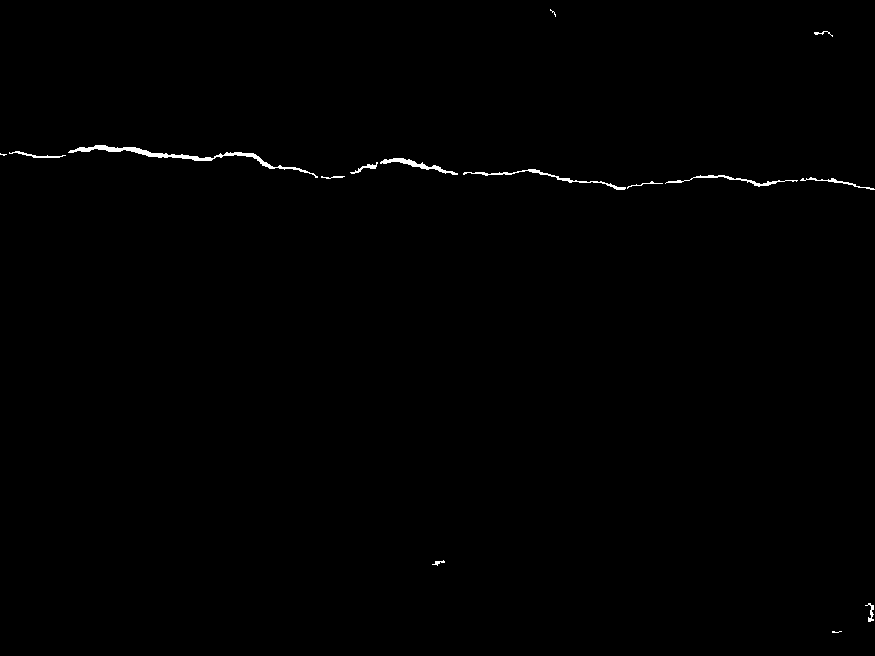}
		\caption{}
	\end{subfigure}%
	\quad
	\begin{subfigure}{.5\textwidth}
		\centering
		\includegraphics[width=160pt]{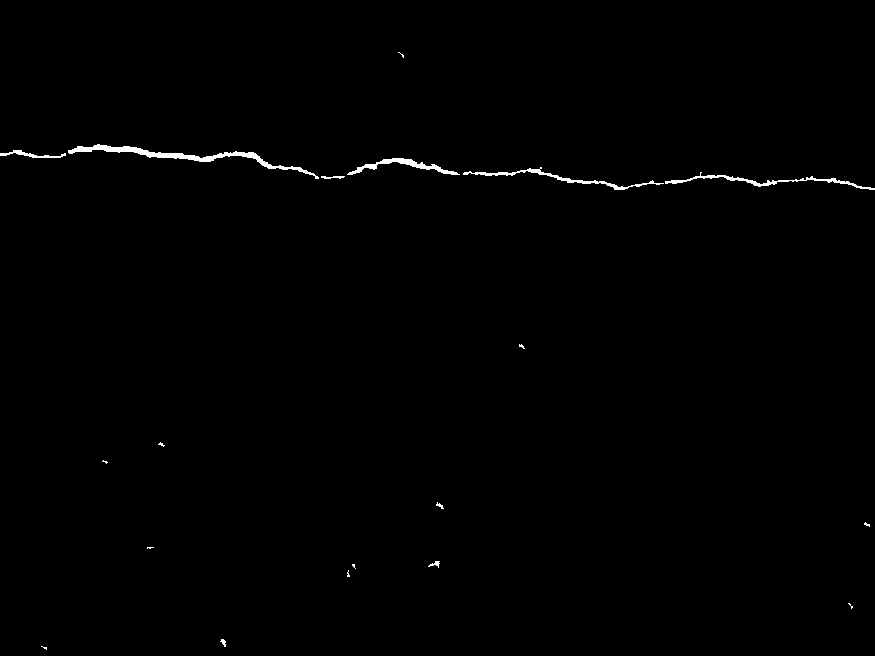}
		\caption{}
	\end{subfigure}
	\begin{subfigure}{.5\textwidth}
		\centering
		\includegraphics[width=160pt]{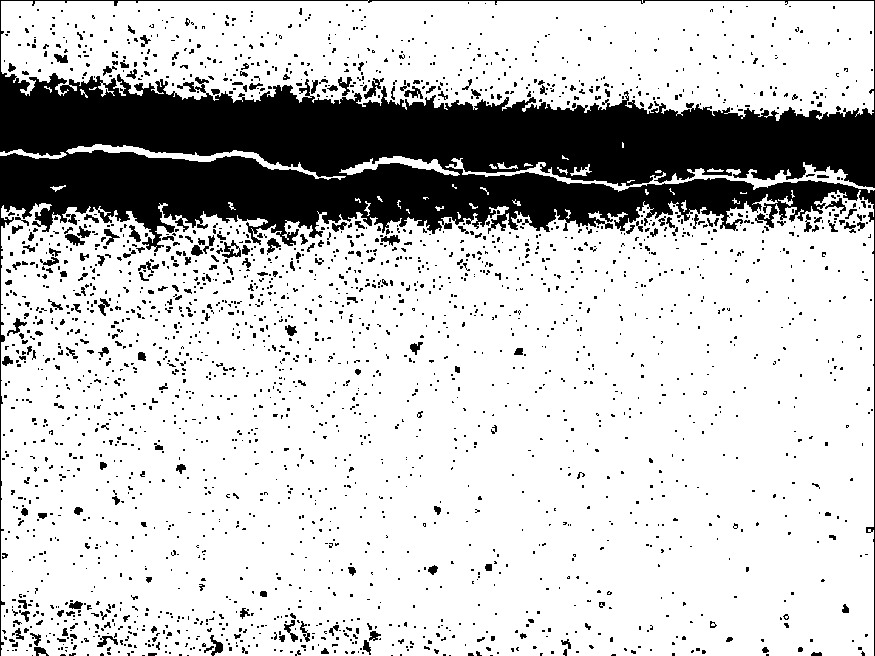}
		\caption{}
	\end{subfigure}
	\caption{Crack detection with segmentation in normal lighting conditions:\\ (a) Surface image, (b) Proposed method, (c) S-K method, and (d) Otsu method}
	\label{fig:R_crack1}
\end{figure}

\begin{figure}
	\centering
	\begin{subfigure}{.5\textwidth}
		\centering
		\includegraphics[width=160pt]{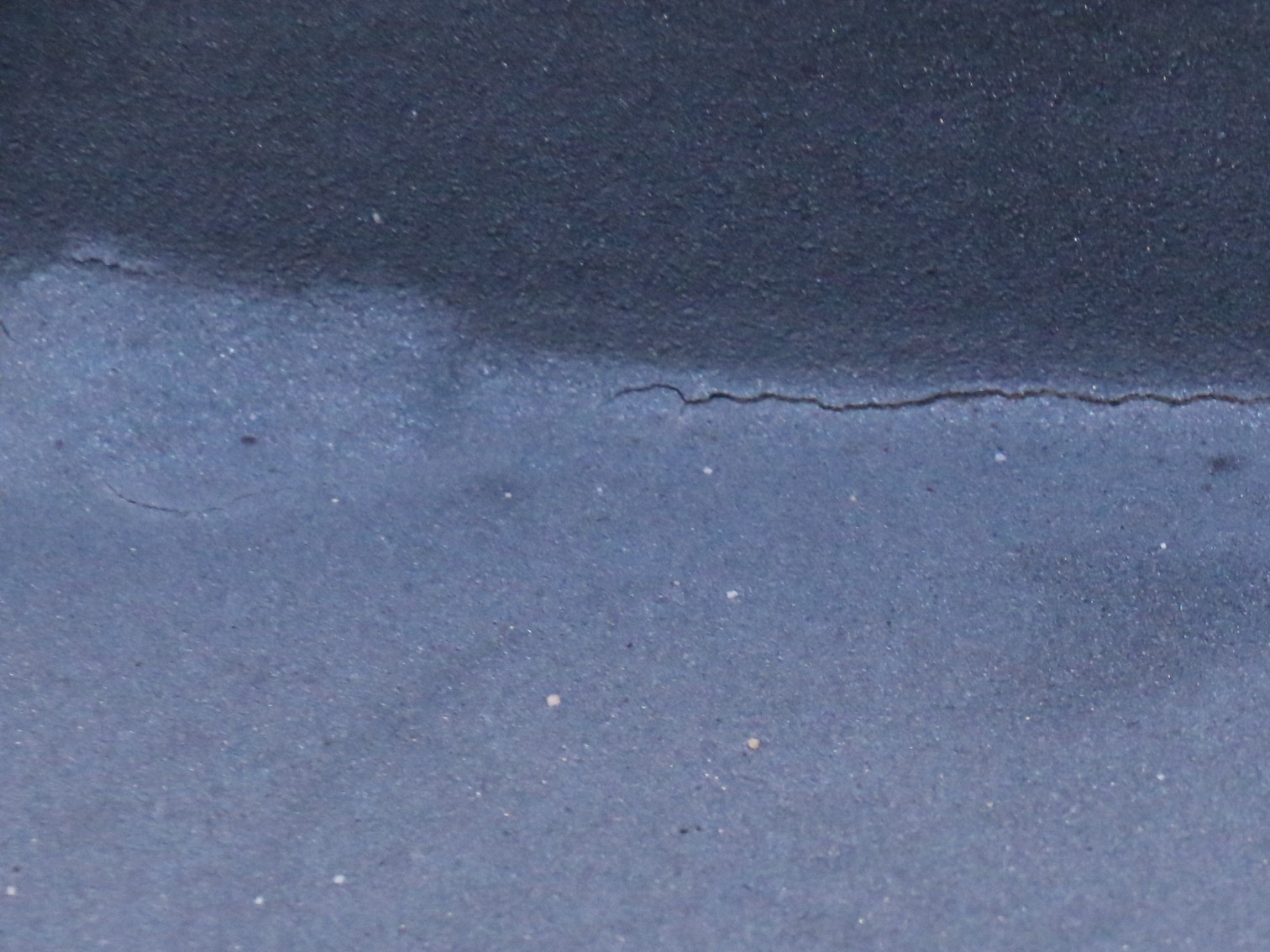}
		\caption{}
	\end{subfigure}%
	\begin{subfigure}{.5\textwidth}
		\centering
		\includegraphics[width=160pt]{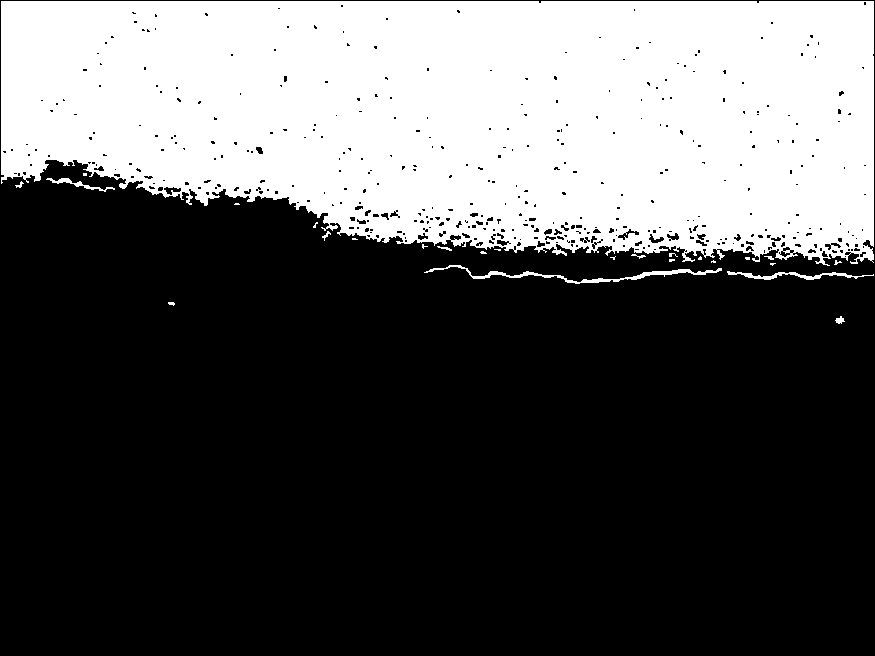}
		\caption{}
	\end{subfigure}%
	\quad
	\begin{subfigure}{.5\textwidth}
		\centering
		\includegraphics[width=160pt]{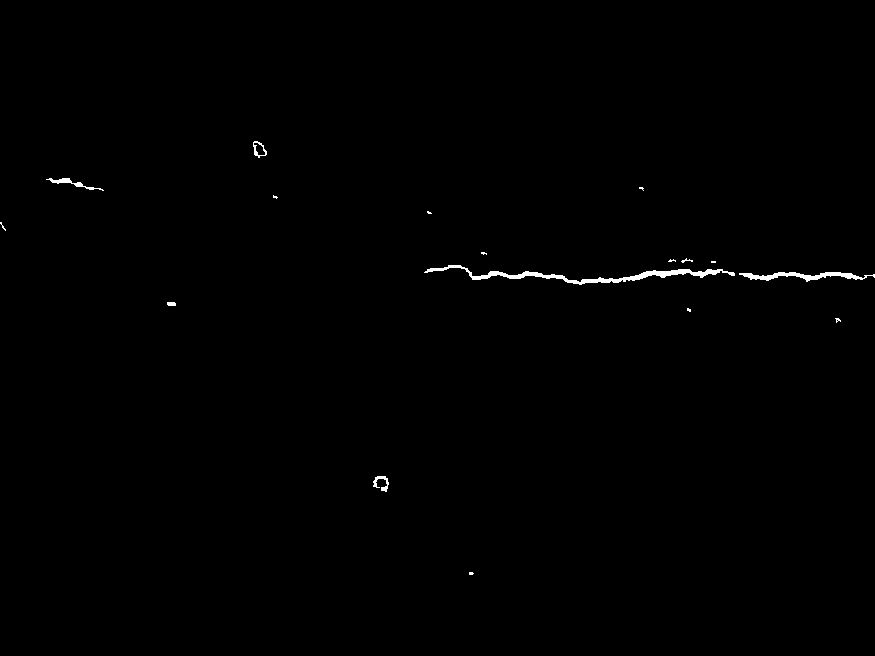}
		\caption{}
	\end{subfigure}
	\quad	
	\begin{subfigure}{.5\textwidth}
		\centering
		\includegraphics[width=160pt]{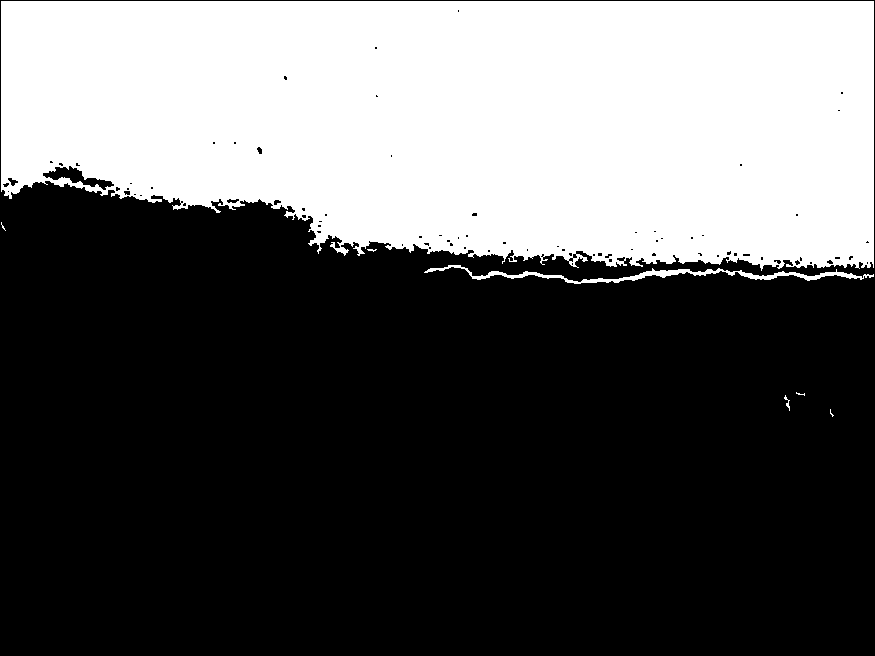}
		\caption{}
	\end{subfigure}
\caption{Crack detection with segmentation in low lighting conditions:\\  (a) Surface image, (b) Proposed method, (c) S-P method, and (d) Otsu method}
	\label{fig:R_crack2}
\end{figure}

Besides, multiple steel structures are combined to form a model bridge to test the data collection process. After navigating a particular distance, the robot stops to capture images of the surface underneath and send to the ground station. Motion of the robot is controlled remotely from ground station while the acquired data are presented in Fig. \ref{fig:Data}. Acquired images are then stitched together to produce an overall image of the steel surface in inspection, as shown in Fig. \ref{fig:stitch}. To this end, results of the steel crack detection algorithm are  presented in Fig. \ref{fig:Crack}. It is significant to see that a small crack shown in Fig. \ref{fig:Crack}(a) was detected as can be seen in Fig. \ref{fig:Crack}(b). The combination of the global thresholding and line emphasizing method can partially extract the crack from the image background but the result is not quite obvious as thin structures were not distinguished and hence the segmented image displayed a disconnection, as shown in Fig. \ref{fig:Crack}(c). An improvement of the result is depicted in Fig. \ref{fig:Crack}(d) where the crack is fully extracted from the background, demonstrating the merit of our proposed enhanced segmentation method. Small connected components and isolated pixels can be removed by morphological operations. Out of 231 collected crack images, the success rate of our proposed algorithm was 93.1\%, where cracks of more than 3 mm width could be accurately detected. Only 15 images taken under poor lighting conditions were wrongly classified as the contrast between the crack and the image background is minimized. Indeed, we have compared our approach with two other popular binarization algorithms, the Otsu method and Sauvola-Pietikinen (S-P) method \cite{Sauvola}, evaluated in accordance with the following criteria when comparing the segmentation results and the groundtruth:
	\begin{itemize}
		\item the precision index (PI):
			\begin{equation}
				PI=\dfrac{TP}{TP+FP},
			\end{equation}
		\item the sensitivity index (SI):
			\begin{equation}
				SI=\dfrac{TP}{TP+FN},
			\end{equation}
		\item the Dice similarity coefficient (DSC): 
			\begin{equation}
				DSC=\dfrac{2TP}{2TP+FP+FN},
			\end{equation}
	\end{itemize}
\noindent where $TP$ and $TN$ are the correctly-detected pixels corresponding respectively to the crack and background objects, $FP$ is the wrongly-detected crack pixels and $FN$ is the crack pixels  missing in the detection process. Under normal lighting and uniform background conditions as shown in Fig. \ref{fig:R_crack1}(a), a crack candidate is fully extracted by using our proposed method and the S-P method with only a few small FPs, as shown in Fig. \ref{fig:R_crack1}(b) and Fig \ref{fig:R_crack1}(c) while the crack segmentation using the Otsu method appears to be affected by  noise as depicted in Fig. \ref{fig:R_crack1}(d). Figure \ref{fig:R_crack2}(a) shows a particular case of low lighting conditions, where the intensity of a part of the background is smaller than that of the crack. In this case, our approach as well as the Otsu method may incorrectly detect the dark part of the background as presented in \ref{fig:R_crack2}(b) and (d). For the sake of comparison, Tab. \ref{tbl:Evaluation} summarizes the PI, SI and DSC of the segmentation results using each method in normal and poor lighting conditions. It can be seen that the proposed approach provides the highest precision level for the crack in the normal lighting condition as shown in Fig. \ref{fig:R_crack1}(a) with a small number of FP pixels.

\begin{table}
	\centering
\caption{Comparison of various methods for crack detection by segmentation}
	\label{tbl:Evaluation}
	\begin{tabular}{|p{1.2cm}|p{1.28cm}|p{1.28cm}|p{1.28cm}|p{1.28cm}|p{1.28cm}|p{1.28cm}|}
		\hline
		\multirow{2}{*}{Method} & \multicolumn{2}{l|}{~~~~~~~~~  PI}                                         & \multicolumn{2}{l|}{~~~~~~~~~ SI}                                       & \multicolumn{2}{l|}{~~~~~~~~ DSC}                                      \\ \cline{2-7} 
		& normal lighting  & low lighting& normal lighting  & low lighting & normal lighting & low lighting \\
		\hline
		Proposed         & 0.9360                        & 0.0079                        & 0.6507                       & 0.7365                       & 0.7677                       & 0.0156                       \\
		\hline
		S-P    & 0.9210                        & 0.7679                               &  0.8555                       &0.8831                              & 0.8871                       &0.8215                              \\
		\hline
		Otsu      & 0.0076                        & 0.0089                        & 0.9734                       & 0.8836                       & 0.0151                       & 0.0177   \\
		\hline                   
	\end{tabular}
\end{table}

\begin{figure}
\centerline{\includegraphics[width=\linewidth]{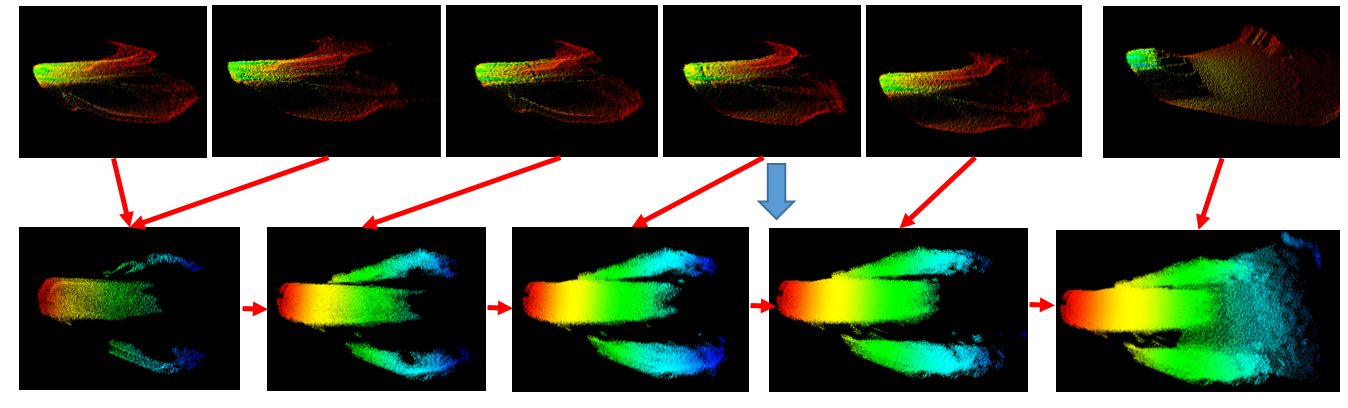}}%
\caption{3D registration and stitching from point cloud data.}
\label{fig:3dreg}
\end{figure}

Regarding the 3D construction capability for navigation planning and further processing purposes, the robot can capture 3D images from ToF camera while moving on the steel surface. The data are saved to the on-board computer's hard disk. Then we apply the registration process which uses the ICP algorithm to align multiple point cloud data. The 3D construction results are presented in Figure \ref{fig:3dreg}.

\section{Conclusion}\label{Sec:Conc}

This paper has presented the development of a steel climbing robot and its data processing system for steel infrastructure monitoring. The robot is capable of carrying multiple sensors for its navigation and for steel surface inspection. The initial prototype is implemented and validated to ensure the robot is able to strongly adhere on steel surfaces in different situations. Captured images are stitched together to create a single image of the steel surface then steel crack detection algorithms are implemented to locate a defect on the stitched image. A 3D map is also generated from images captured by the robot. Thus, in addition to various sensors being integrated for navigation and surface inspection, and the collected visual and 3D images can be transferred to the ground station for further monitoring purposes. Further work will include localization using combined odometry and camera data, improvement of the map construction process as well as detection algorithms from stitched images. Additionally, a cooperative process can be used to employ multiple robots for inspecting and monitoring large-scale steel infrastructure.

\section*{Acknowledgements}
This work is partially supported by the INSPIRE University Transportation Center provided by the U.S. Department of Transportation Office of the Assistant Secretary for Research and Technology (USDOT/OST-R) under grant agreement No. 69A3551747126 and by participating state departments of transportation; the University of Nevada, Reno; the National Science Foundation under the grant  NSF-IIP \#1559942; and the 2016 Data Arena grant from the University of Technology Sydney. The second author would like to acknowledge support from an Australian Awards scholarship.


\end{document}